\RequirePackage{iftex}
\ifPDFTeX\pdfoutput=1\fi
\documentclass[11pt,a4paper,google]{google}
\usepackage{iftex}
\ifPDFTeX\else
  \microtypesetup{tracking=false}
  \usepackage{fontspec}
\fi

\usepackage[authoryear,sort&compress,round]{natbib}
\usepackage{array,makecell,adjustbox,multirow}
\usepackage{fvextra}
\usepackage{placeins}
\usepackage{float}
\usepackage{wrapfig}
\usepackage{etoolbox}

\usepackage{amsmath,amsfonts,bm}

\def\eqref#1{equation~\ref{#1}}

\def\1{\bm{1}}

\DeclareMathAlphabet{\mathsfit}{\encodingdefault}{\sfdefault}{m}{sl}
\SetMathAlphabet{\mathsfit}{bold}{\encodingdefault}{\sfdefault}{bx}{n}

\DeclareRobustCommand{\paperchange}[1]{{\color{black}#1}}
\definecolor{GoogleBlue}{HTML}{4285F4}
\definecolor{GoogleRed}{HTML}{EA4335}
\definecolor{GoogleYellow}{HTML}{FBBC05}
\definecolor{GoogleGreen}{HTML}{34A853}

\DeclareRobustCommand{\BenchName}{\textit{%
  \VISPrefix-\textcolor{GoogleBlue}{B}\textcolor{GoogleGreen}{e}%
  \textcolor{GoogleRed}{n}\textcolor{GoogleBlue}{c}\textcolor{GoogleRed}{h}}}
\DeclareRobustCommand{\MethodName}{\textit{\textbf{%
  \VISPrefix-\textcolor{GoogleBlue}{G}\textcolor{GoogleGreen}{r}%
  \textcolor{GoogleRed}{o}\textcolor{GoogleBlue}{u}%
  \textcolor{GoogleRed}{n}\textcolor{GoogleYellow}{d}}}}
\DeclareRobustCommand{\appref}[1]{\hyperref[#1]{Appendix~\ref*{#1}}}
\usepackage{amsmath,amssymb}
\usepackage{booktabs,array}
\usepackage{graphicx,makecell}
\usepackage{enumitem}
\usepackage{xcolor}
\usepackage{hyperref}
\providecommand{\BenchName}{\textit{VIS-Bench}}
\providecommand{\MethodName}{\textit{VIS-Ground}}

\usepackage{fvextra}

\hypersetup{colorlinks=true,allcolors=black,
  pdftitle={VIS-Ground: Video Interactive Storytelling with Contextual Grounding},
  pdfauthor={Bingxuan Li, Yiwen Song, Xueqing Wu, Yanzhou Pan, Yang Li, Kuang Su, Jingyun Liu, Sebastian Ko, Huan Zhang, Tong Zhang, Nanyun Peng, Tomas Pfister, Yale Song}}
\title{\MethodName{}: Video Interactive Storytelling with Contextual Grounding}
\uselogo{}
\reportnumber{}
\correspondingauthor{}
\renewcommand{\today}{}

\author[1,2,*,\textdagger]{Bingxuan Li}
\author[1,\textdagger]{Yiwen Song}
\author[3]{Xueqing Wu}
\author[1]{Yanzhou Pan}
\author[1]{Yang Li}
\author[1]{Kuang Su}
\author[1]{Jingyun Liu}
\author[1]{Sebastian Ko}
\author[2]{Huan Zhang}
\author[2]{Tong Zhang}
\author[1]{Nanyun Peng}
\author[1]{Tomas Pfister}
\author[1,\textdagger]{Yale Song}
\affil[1]{Google}
\affil[2]{University of Illinois at Urbana-Champaign}
\affil[3]{University of California, Los Angeles}

\begin{abstract}
Video interactive storytelling enables viewers to actively steer how a video unfolds. However, once we allow viewers to intervene during generation, a new challenge arises: The viewer's request can have latent dependencies on both the grounding source and the current rendered video state. These dependencies may not be explicitly stated in any individual input, but emerge only when the source, rendered history, and new viewer intent are considered jointly. Existing interactive video generation systems primarily emphasize following viewer instructions, while source-grounded video generation methods focus on aligning generated content with an external narrative or knowledge source. This leaves a fundamental question underexplored: What context should a generation model ground on during interactive continuation, and how can heterogeneous, unstructured inputs be transformed into such grounding context? In this work, we formulate contextual grounding as the process of transforming heterogeneous input context into an executable constraint model for video generation. To address this challenge, we introduce \MethodName{}, which performs Structured Context Abstraction to recover grounded states and cross-context dependencies, Generation Constraints Induction to project relevant dependencies into candidate-specific constraints, and Constrained Video Generation to enforce these constraints through planning, verification, revision, and rendering. Across three video generation backbones, \MethodName{} consistently achieves the highest overall composite score, reaching an average absolute improvement of 10.3 points over the strongest per-backbone baselines. Detailed analysis further shows gains across both narrative and knowledge grounding, and reveals remaining challenges in dependency extraction, and faithful realization during video rendering.
\end{abstract}

\DeclareRobustCommand{\projectwebsitelink}{%
  \href{https://vis-ground.github.io}{%
    \raisebox{-0.1em}{\includegraphics[height=0.85em]{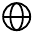}}\kern0.3em\textbf{Project Page:} \nolinkurl{https://vis-ground.github.io}}}

\makeatletter
\renewcommand{\maketitle}{%
  \begingroup
  \setlength{\parindent}{0pt}%
  \setlength{\parskip}{0pt}%
  {\raggedright\titlefont\@title\par}%
  \vspace{7pt}%
  {\raggedright\@author\par}%
  \begingroup
  \renewcommand{\thefootnote}{\fnsymbol{footnote}}
  \footnotetext[1]{Work done as Student Researcher at Google.\quad\textsuperscript{\textdagger}Corresponding authors.}
  \endgroup
  \vspace{7pt}%
  {\fontsize{9}{11}\selectfont\projectwebsitelink{}\par}%
  \vspace{5pt}%
  \begingroup
\setlength{\intextsep}{0pt}
\captionsetup{font=footnotesize,skip=4pt}
\begin{figure}[H]
    \centering
    \includegraphics[width=0.86\linewidth]{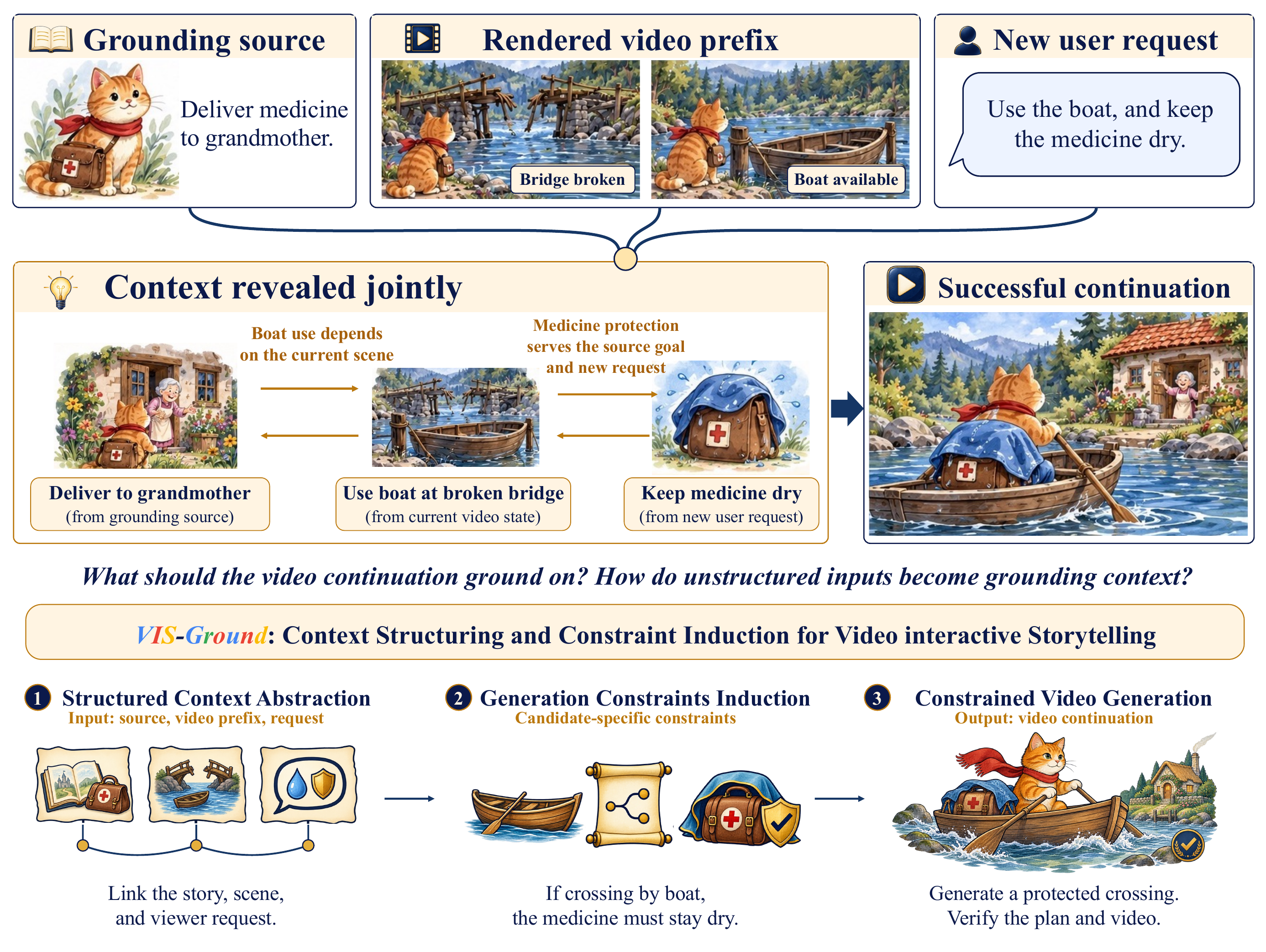}
    \caption{\textbf{Video interactive storytelling with contextual grounding.} A viewer's request can depend jointly on the grounding source and the rendered video prefix. These dependencies become generation constraints when projected onto a candidate future continuation. \MethodName{} structures the context, induces candidate-specific constraints, and generates and verifies the video continuation.}
    \label{fig:teaser}
\end{figure}
\endgroup
  \vspace{3pt}%
  {\noindent\textbf{Abstract}\par\vspace{8pt}%
   \noindent\parbox{\linewidth}{\mdseries\linespread{1}\fontsize{10}{11.5}\selectfont\theabstract}\par}%
  \vspace{6pt}%
  {\fontsize{9}{11}\selectfont\itshape Keywords: \@keywords\par}%
  \endgroup
  \thispagestyle{firststyle}%
}
\makeatother

\begin{document}
\newgeometry{left=2.2cm,right=2.2cm,top=2cm,bottom=2cm}
\maketitle
\restoregeometry
\section{Introduction}

Video storytelling communicates narratives and explains complex ideas through language, imagery, motion, and sound. Yet a viewer's interests are rarely fixed before watching begins: A narrative choice may invite questions about consequences, or an unfamiliar concept may prompt a need for concrete examples. Such requests arise from the content just observed, motivating storytelling systems that allow viewers to shape how a video continues. While human--AI writing tools like Wordcraft demonstrate the value of interaction during story creation~\citep{yuan2022wordcraft}, this remains underexplored in video storytelling. In our formative study of knowledge-grounded video storytelling, the reported mean pre-to-post gains were 47.9 percentage points with interaction and 32.1 without interaction. These results motivate further study of interaction as a way to support viewers' understanding.

Supporting such interaction, however, is not simply a matter of appending a new prompt to a video generator. Once part of a video has been rendered, it establishes concrete visual states, events, and information that subsequent generations must respect. At the same time, an underlying narrative or factual source may impose additional dependencies that are not directly visible in the video. A viewer request therefore cannot be interpreted in isolation: some aspects of the continuation may adapt freely, while others are constrained by what has already happened or by what the grounding material supports. Direct generation from the source, video prefix, and latest request leaves these relationships implicit, making it easy to introduce incompatible state changes, contradict previously communicated information, or satisfy the request in a way unsupported by the source. The central challenge is therefore not only \emph{what to generate next}, but \emph{how to determine which contextual information matters and how it should constrain the continuation}.

We formulate this problem as \textbf{Video Interactive Storytelling with Contextual Grounding}. Given a grounding source (e.g., a narrative, paper, wiki page, or reference document), a previously rendered video prefix, and a new viewer request, the system generates a continuation that responds to the viewer while remaining compatible with the established context. To address the viewer's new request, it requires jointly reasoning over three heterogeneous forms of information: the grounding source specifies what is supported, the video prefix records what the viewer has actually observed, and the evolving story captures how events and information relate over time.

Existing work provides important pieces of this capability, but does not directly address their combination. Recent video storytelling systems develop increasingly capable pipelines for story planning, multimodal production, and coherent long-form generation~\citep{xu2025mmstoryagent,wu2025movieagent,zheng2024videogenofthought,song2026codirector,vimax2026}. Their primary formulations, however, generate videos from an initially specified story, script, or creative intent. Complementary work on interactive video generation enables generation to evolve with changing text prompts~\citep{yin2025causvid,kodaira2025streamdit,yang2025longlive,streamdiffusionv2}, while interactive world models support action-conditioned exploration and controllable world events~\citep{bruce2024genie,oasis2024,matrixgame2025,huang2025vid2world0,deepmind2025genie3}. These advances make video generation increasingly responsive, but responsiveness alone does not determine whether a newly requested continuation is compatible with the source and with what has already been shown. What is missing is an mechanism for transforming these heterogeneous inputs into a grounded representation of the current context and reasoning about the consequences they impose on future generation.

Motivated by this intuition, we introduce \MethodName{}, a framework that separates \emph{what the available context establishes} from \emph{what its relationships imply for generation}. \MethodName{} first performs \textbf{Structured Context Abstraction}, transforming them into aligned video, story, and source states that make relevant entities, events, claims, state transitions, and cross-input relationships explicit. It then performs \textbf{Generation Constraints Induction}, reasoning over this structure to derive instance-specific conditions that a valid continuation should satisfy. These grounded constraints provide an explicit interface between contextual understanding and generation: they guide script planning, identify unsupported or conflicting proposals during verification, and provide targeted feedback for revision. \paperchange{Finally, \textbf{Constrained Video Generation} applies contextual verification both before and after rendering through an iterative plan--verify--revise--render process, because a valid script does not guarantee that a video generator faithfully realizes it.}

To systematically study this problem and address the absence of dedicated benchmarks for the task, we further introduce \BenchName{} \paperchange{(Video Interactive Storytelling)}, a benchmark pairing grounding sources, observed video prefixes, and viewer interventions across narrative- and knowledge-grounded settings. Our evaluation measures source faithfulness, story continuity, and interaction fulfillment individually and jointly, together with visual consistency and video quality, allowing us to distinguish contextual grounding from perceptual generation quality. Across three different video generation backbones (Omni-1.1-Flash~\citep{google2026omni11}, Veo~3.1~\citep{deepmind2025veo31}, and MiniMax-H3~\citep{minimax2026h3}) \MethodName{} {achieves the highest overall composite score with each backbone. The gains over the strongest corresponding baselines are 12.3 points for Omni, 15.3 for Veo, and 3.4 for MiniMax.}


\Needspace{24\baselineskip}
\section{Video Interactive Storytelling with Contextual Grounding}
\label{sec:task-context}

\subsection{Formative Study}
\label{sec:preliminary-study}

\begin{wrapfigure}{r}{0.52\textwidth}
\centering

\includegraphics[width=\linewidth]{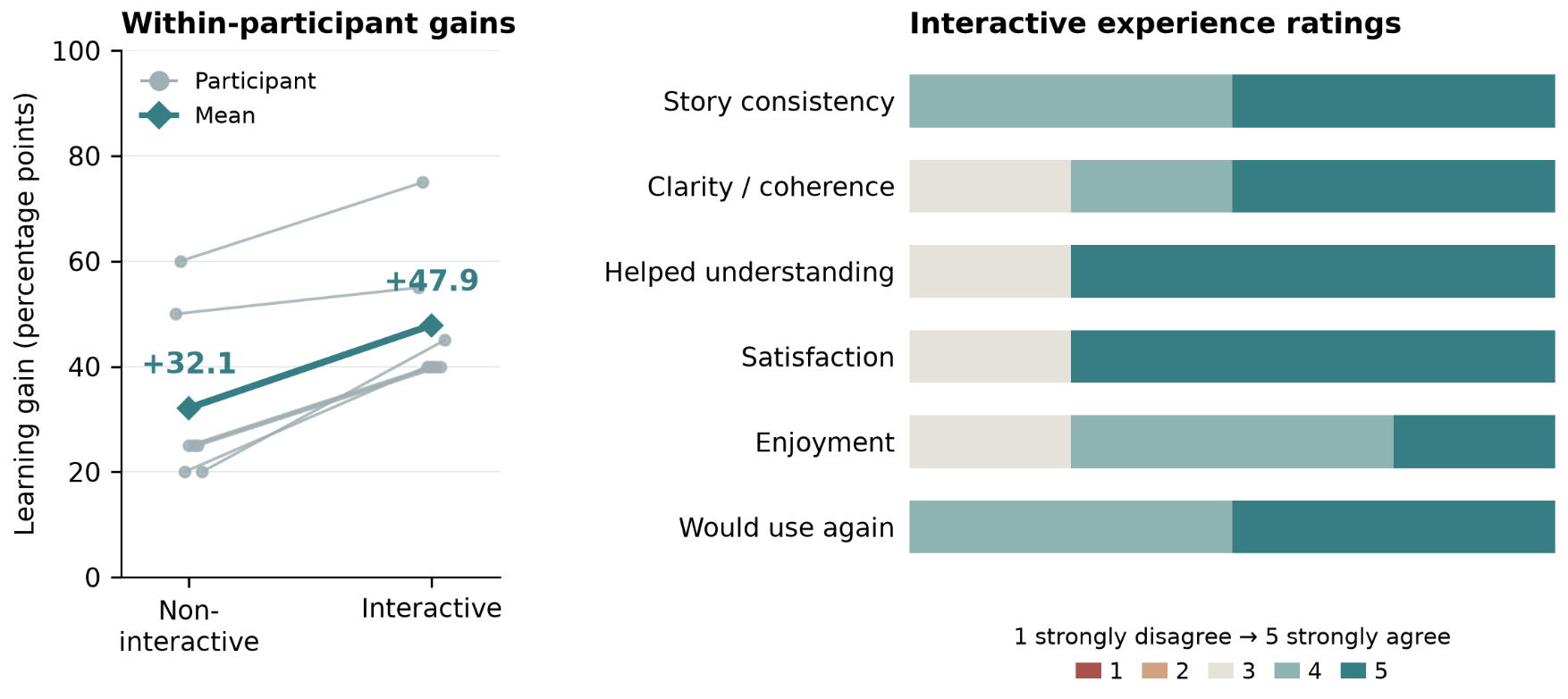}

\caption{{\textbf{Formative Study Results.} The study enrolled nine volunteers. \textit{Left:} Learning gains for the seven participants who passed the attention test. \textit{Right:} Experience ratings from four respondents.}}
\label{fig:preliminary-study}

\end{wrapfigure}

Prior work on human--AI co-creation suggests that viewer intent often evolves after generated content becomes visible. Analyses of large-scale human--AI interaction logs find that viewers revise, extend, or reformulate their intentions after observing generated outputs in 18--30\% of sessions \citep{mysore2025prototypical}.

\leavevmode{We conducted a formative study with nine volunteers to examine interaction in knowledge-based video storytelling. Figure~\ref{fig:preliminary-study} reports learning gains for the seven participants who passed the attention test. Among these seven participants, mean gains were higher with interaction; the experience ratings were concentrated toward the positive end of the five-point scale. These exploratory observations motivate studying how viewers can ask questions and redirect a continuation while preserving its grounding source and evolving video context. They do not establish a learning-gain estimate for all nine participants. See \appref{app:formative} for details.}

\subsection{Task Formulation}
\label{sec:task-formulation}

Given a grounding source $\mathcal{G}$, a previously generated video prefix $v_{\mathrm{prefix}}$, and a current viewer request $u$, the model generates a video continuation $v_{\mathrm{cont}}$:
$v_{\mathrm{cont}} \sim
p_\theta\!\left(
v_{\mathrm{cont}}
\mid
\mathcal{G}, v_{\mathrm{prefix}}, u
\right)$
Here, $\mathcal{G}$ provides persistent context the generated story should respect, and $v_{\mathrm{prefix}}$ provides the visual and narrative context for the continuation. Rather than explicitly conditioning on all the previous viewer instructions, each continuation relies on $v_{\mathrm{prefix}}$ as a sufficient record of earlier instructions.




\section{Method}
\label{sec:method}

We introduce \MethodName{}, a framework for video interactive storytelling that treats contextual grounding as a \emph{context compilation} problem.

\begin{figure}[h]
    \centering

    \includegraphics[width=0.95\linewidth]{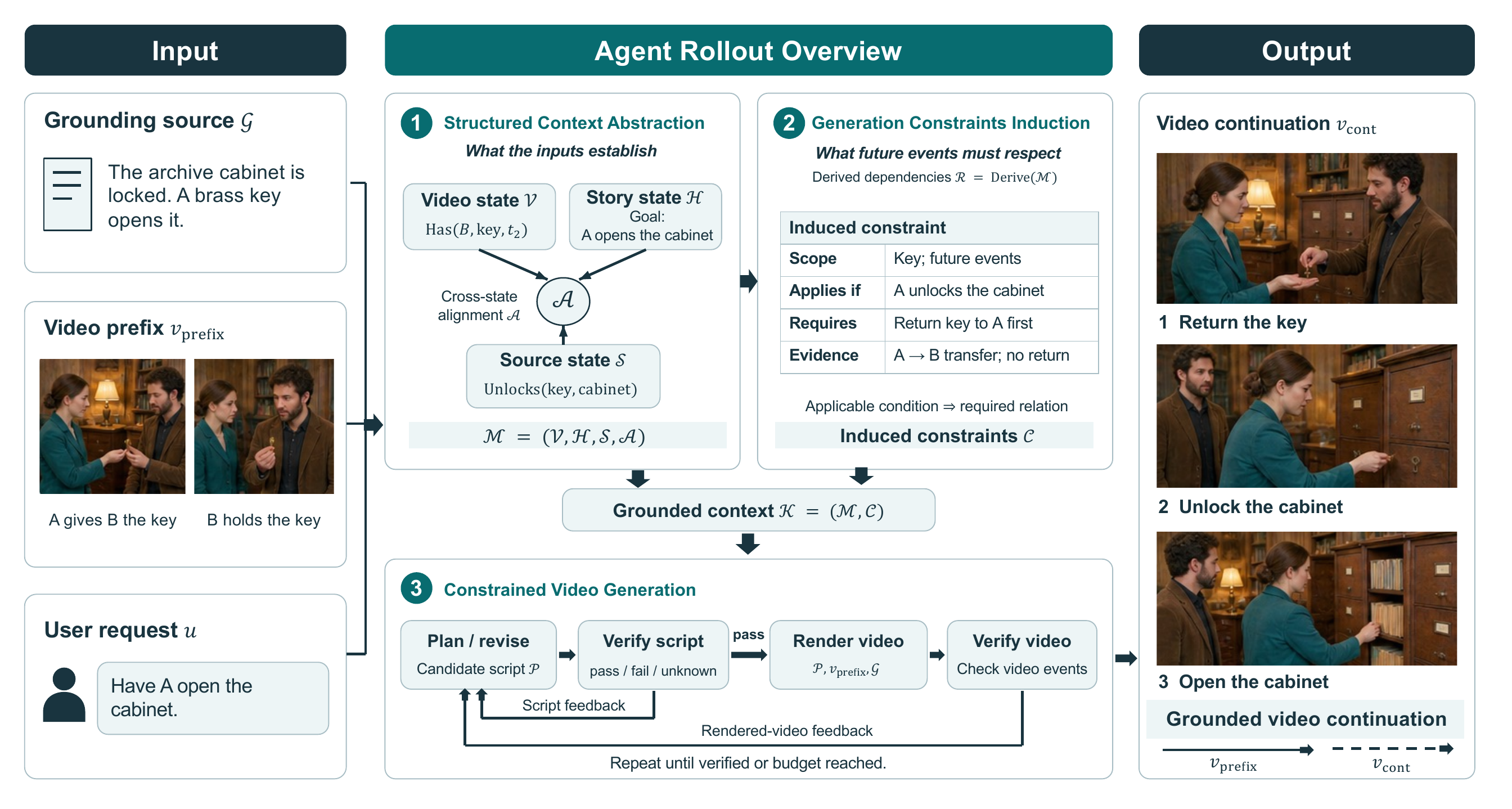}

    \caption{\paperchange{\textbf{Overview of \MethodName{}.} Structured Context Abstraction aligns the video, story, and source states into $\mathcal{M}$. Generation Constraints Induction derives dependencies $\mathcal{R}$ from $\mathcal{M}$ and activates candidate-specific constraints $\mathcal{C}$. Constrained Video Generation uses script and video verification to repair the continuation $v_{\mathrm{cont}}$ from the prefix $v_{\mathrm{prefix}}$ and the viewer request.}}
    \label{fig:method}

\end{figure}

Our central idea is to compile heterogeneous context into a \emph{grounded dependency model} and use this model to control subsequent generation. As illustrated in Figure~\ref{fig:method}, \MethodName{} first converts the raw inputs into grounded facts and cross-context dependencies (Sec.~\ref{sec:dependency-compilation}). These dependencies are then projected onto a candidate continuation, where only the relations made relevant by the proposed future are instantiated as generation constraints (Sec.~\ref{sec:constraint-projection}). Finally, the resulting constraints guide planning, script repair, rendering, and output-level correction (Sec.~\ref{sec:dependency-generation}). \appref{app:method} expands the representation and repair procedure, with an illustrative dependency trace in \appref{app:trace}.

\subsection{Compiling Context into Grounded Dependencies}
\label{sec:dependency-compilation}

The grounding source, video prefix, and viewer request encode different views of an evolving situation. The source may establish factual or narrative relations, the rendered prefix determines the current visible state, and the viewer request specifies how the story should continue.

The challenge is not merely to summarize each input, but to recover the dependencies that emerge from their interaction.

\noindent\textbf{Grounded State Extraction.}
We first transform each input into a structured set of grounded facts. We extract entities, attributes, events, relations, and state transitions using typed predicates with arguments, temporal scopes, and evidence links with VLMs. We then organize these predicates into three complementary states:

\begin{itemize}[leftmargin=*, noitemsep, topsep=0pt]

\item \textbf{Video state $\mathcal{V}$} represents what has been visually established in the rendered prefix, including entities, attributes, locations, possession, actions, and state transitions. Temporal annotations distinguish prior observations from the latest established state.

\item \textbf{Story state $\mathcal{H}$} represents the evolving narrative state, including completed events, active goals, unresolved relations, causal dependencies, and information already conveyed.

\item \textbf{Source state $\mathcal{S}$} represents atomic claims, events, definitions, and relations supported by the grounding source, while preserving qualifications, conditions, and supporting evidence.

\end{itemize}


\noindent\textbf{Cross-Context Dependency Construction.}
The aligned states are compiled into a grounded dependency model $\mathcal{M}=(\mathcal{V},\mathcal{H},\mathcal{S},\mathcal{A})$, \paperchange{where $\mathcal{A}$ contains cross-context alignments. We derive a dependency set $\mathcal{R}=\operatorname{Derive}(\mathcal{M})$ from the aligned predicates; $\mathcal{R}$ is a derived object rather than an additional component of $\mathcal{M}$.} Each dependency expresses how one established fact constrains another event or state. These relations include temporal precedence, causal prerequisites, state-transition requirements, persistence relations, compatibility conditions, and source-supported dependencies. Importantly, many such dependencies are not explicitly stated in any individual input. They arise only after the inputs are jointly grounded.



\subsection{Projecting Dependencies into Generation Constraints}
\label{sec:constraint-projection}

\paperchange{We induce constraints from the dependency set $\mathcal{R}$ derived from the grounded model $\mathcal{M}$, activating those relevant to each candidate continuation.} These dependencies link the source, rendered prefix, and viewer request. Our method does not directly optimize final-video evaluation metrics: their criteria are neither optimization objectives nor predefined categories for constraint induction.

\noindent\textbf{Candidate-Conditioned Activation.}
Let $\mathcal{P}^{(k)}$ denote a candidate continuation script. For each dependency, we determine whether the candidate introduces an event, relation, or state transition for which that dependency becomes relevant. An activated dependency is represented as a generation constraint
$c_j=(s_j,a_j,b_j,e_j)$,
where $s_j$ specifies its entity and temporal scope, $a_j$ is an applicability condition, $b_j$ is the relation that must hold, and $e_j$ contains the grounded evidence supporting the dependency. The constraint is activated when $a_j(\mathcal{P}^{(k)},\mathcal{M}) = 1$, and once activated, requires $b_j(\mathcal{P}^{(k)},\mathcal{M})$ to hold within scope $s_j$.


This formulation makes the constraints \emph{candidate-conditioned}. A possession dependency, for example, need not constrain a continuation unless the proposed script introduces an action that requires a particular character to possess the corresponding object. Similarly, a source-defined prerequisite is activated only when the candidate attempts an event whose validity depends on that prerequisite.


\noindent\textbf{Constraint Evaluation.}
For an activated constraint $c_j$, we compare the relations implied by the candidate script with the grounded dependency model. Each constraint evaluation returns $\{\mathrm{pass},\mathrm{fail},\mathrm{unknown}\}$. A \texttt{pass} indicates that the proposed continuation is compatible with the dependency; a \texttt{fail} identifies a conflicting event, state transition, or missing prerequisite; and an \texttt{unknown} indicates that the available evidence is insufficient to determine compatibility. Because every constraint retains links to its supporting predicates and evidence, failures can be traced back to the contextual dependency that caused them and used to produce targeted revision feedback.

\subsection{\paperchange{Constrained Video Generation}}
\label{sec:dependency-generation}

The grounded dependency model is not used merely as additional conditioning for the video generator. Instead, it remains active throughout generation and provides a common basis for proposing, checking, and repairing the continuation.

\noindent\textbf{Planning and Script Repair.}
The planner first proposes a candidate script conditioned on the grounded context,
$\mathcal{P}^{(0)}
\sim
p_{\theta}
(\mathcal{P}\mid\mathcal{M},u).$
The dependencies activated by $\mathcal{P}^{(0)}$ are then instantiated as constraints and evaluated. When a constraint fails, the framework identifies both the conflicting portion of the script and the grounded dependency that it violates. This produces targeted feedback for revision rather than regenerating the continuation from scratch.

The revised script is therefore generated as
$
\mathcal{P}^{(k+1)}
=
\operatorname{Revise}_{\theta}
\bigl(
\mathcal{P}^{(k)},
\mathcal{M},
F^{(k)}
\bigr),
$
where $F^{(k)}$ contains the failed or unresolved dependencies and their supporting evidence. Because revision may introduce new events or state transitions, constraint activation is recomputed after each update. Planning and verification therefore form an iterative dependency-guided repair process in which the relevant obligations evolve together with the candidate continuation.

\noindent\textbf{Rendering and Output Repair.}
Once a candidate script $\mathcal{P}^{*}$ satisfies the activated script-level constraints, it is passed to the video generator: \paperchange{$v_{\mathrm{cont}} \sim p_{\psi}(v_{\mathrm{cont}}\mid\mathcal{P}^{*},v_{\mathrm{prefix}},\mathcal{G}).$}
Script-level validity, however, does not guarantee that the rendered video realizes the intended events or state transitions. We therefore extract the realized relations from \paperchange{$v_{\mathrm{cont}}$} and project the same grounded dependencies onto the generated output. If the video fails to realize an activated dependency---for example, an object transfer is omitted, a persistent attribute changes, or a prerequisite event is visually absent---the corresponding evidence is converted into output-level feedback. The planner then repairs the script or makes the missing dependency more explicit before rendering again. In this way, \MethodName{} uses a single grounded dependency model across both reasoning and generation. Context is first compiled into dependencies, dependencies are instantiated as constraints only when a proposed future makes them relevant, and violations are used to repair either the planned continuation or its rendered realization. The process terminates when the generated continuation satisfies the applicable grounded dependencies or the computation budget is exhausted.

\section{Experiments}
\label{sec:experiments}


\subsection{\BenchName{}: \paperchange{Video Interactive Storytelling Benchmark}}
\label{sec:benchmark}

\begin{figure}[h]
    \centering
    \includegraphics[width=1\linewidth]{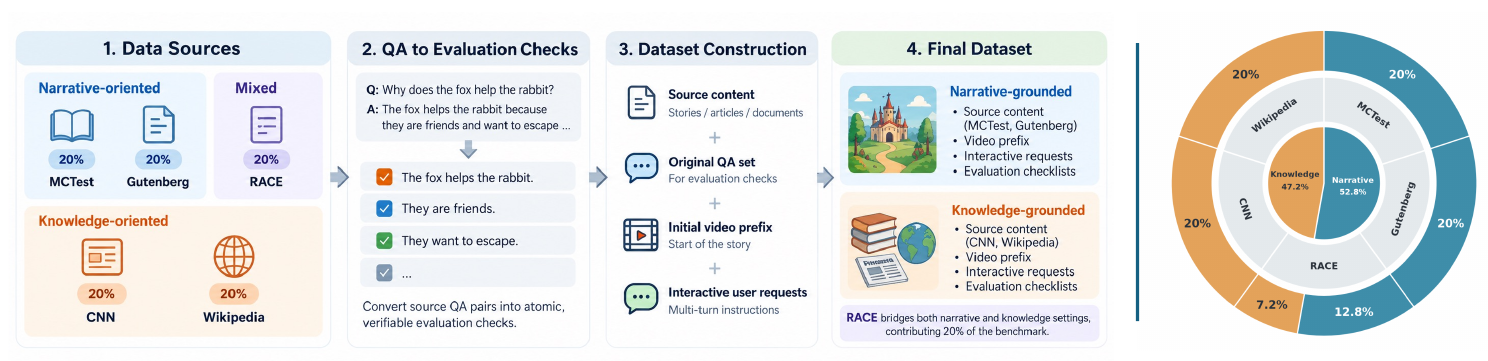}

    \caption{Dataset Construction and Dataset Statistics}

    \label{fig:dataset}
\end{figure}

To systematically study this problem and address the absence of dedicated benchmarks for the task, we introduce \BenchName{}. It considers two grounding settings:
\begin{itemize}[leftmargin=*, noitemsep, topsep=0pt]
    \item \textbf{Narrative grounding} uses a core narrative, such as
    a story outline, script, or plot description, that specifies
    important characters, events, relationships, and intended story
    progression.
    \item \textbf{Knowledge grounding} uses an external information
    source, such as a scientific paper, Wikipedia article, or document,
    whose factual content should be faithfully communicated through
    the story.
\end{itemize}

We construct the dataset from CoQA \citep{coqa}. We sample 250 source passages from MCTest, Gutenberg, RACE, CNN, and Wikipedia, covering 132 narrative- grounded and 118 knowledge-grounded sources. Each source passage is converted into a scene-based story with aligned source facts and rendered in two visual styles: cinematic realism and cutout animation. We generate 1,004 interaction instructions at designated interruption points, yielding 2,008 planned instruction–style instances. All instances are assigned to evaluation; no training split is used. Each sample contains the source passage, video prefix, viewer instruction, and relevant reference assets, with target facts and continuation requirements reserved for evaluation. Generated videos undergo quality checks and repair, instruction alignment and information leakage are audited before packaging. See \appref{app:benchmark} for more details.


\subsection{Evaluation Metrics}
\label{sec:bench_metrics}

We evaluate each generated video along six dimensions. All metrics are normalized
to a $0$--$100$ scale, with higher values indicating better performance.
\appref{app:rubric} gives the rating anchors, \appref{app:aggregation} specifies normalization and judge aggregation

\noindent\textbf{Semantic Metrics.}
Semantic evaluation is based on metric-specific checklists. Each applicable
check is rated on a five-level scale corresponding to
\emph{absent or contradictory}, \emph{mostly unsuccessful},
\emph{partially fulfilled}, \emph{mostly fulfilled}, and
\emph{fully fulfilled}.

\begin{itemize}[leftmargin=*, noitemsep, topsep=0pt]
    \item \textbf{Source Faithfulness (SF)}
    measures whether the generated video remains faithful to the  grounding source $\mathcal{G}$. Its checklist is primarily derived from the original source QA annotations and captures source-specific facts, events, relationships, causal dependencies, and other persistent constraints.

    \item \textbf{Story Continuity (SC)} measures whether the rendered continuation is consistent with the video prefix. The evaluation considers established events, object states, character identities and relationships, spatial configurations, and causal history.

    \item \textbf{Interaction Fulfillment (IF) } measures how completely the continuation satisfies the latest viewer instruction. Each instruction is decomposed into atomic requirements,
    including, where applicable, requested subjects, actions, concepts,
    explanations, and presentation requirements.

    \item \textbf{Joint Constraint Satisfaction (JCS)} captures the model's overall ability to satisfy the three semantic requirements jointly. It is computed as the equal-weight average of SF, SC, and IF.
\end{itemize}

\noindent\textbf{Perceptual Metrics.}
In addition to semantic correctness, we evaluate whether the generated
video remains visually coherent and maintains high perceptual quality.

\begin{itemize}[leftmargin=*, noitemsep, topsep=0pt]
    \item \textbf{Visual Consistency (VC)} measures whether recurring characters, objects, environments, and visual styles remain coherent across the video. The evaluator considers
    both consistency between the existing video prefix and the generated video and consistency within the generated video itself.
    \item \textbf{Video Quality (VQ)} measures the perceptual quality of the generated video,
    using sampled-frame image integrity and composition; these scores do not assess native motion or audio quality. VQ evaluates generation quality independently of source or instruction compliance.
\end{itemize}

\subsubsection{Evaluation Protocol}
\label{sec:benchmark_protocol}

\noindent\textbf{Auto-Evaluator.} We use \texttt{Gemini~3.8 Flash}~\citep{google2026gemini38} and \texttt{Claude Opus 5}~\citep{anthropic2026opus5} as two independent auto-evaluators. We conduct blinded human evaluation to validate the evaluation rubrics and measure agreement between human judgments and the automatic evaluators. For each output, each judge independently produces per-check ratings together with supporting evidence, we then compute each metric independently for the two judges and average their scores. For the overall benchmark composite, we give the two grounding settings equal weight. \appref{app:evaluation} expands the evaluation rubric and score definitions; \appref{app:human-validation} describes the human-validation protocol and agreement results.


\noindent\textbf{Baselines.}
Interactive streaming models such as StreamDiT and LongLive~\citep{kodaira2025streamdit,yang2025longlive} emphasize low-latency, prompt-driven generation. Our task requires source-grounded, 48-second multi-shot continuation from a supplied prefix. We select four complementary baselines:
\textit{Direct Generation} tests the backbone's native capabilities;
\textit{MM-StoryAgent}~\citep{xu2025mmstoryagent} uses multi-agent storytelling;
\textit{MovieAgent}~\citep{wu2025movieagent} uses hierarchical long-form planning; and
\textit{VideoGen-of-Thought}~\citep{zheng2024videogenofthought} uses explicit intermediate reasoning.
Although not originally designed for interactive continuation, these baselines test whether direct generation, agent coordination, and structured planning support our task under shared inputs and backbones, isolating contextual reasoning from synthesis capability. \appref{app:baselines} details the adaptations.

\noindent\textbf{Video Backbones.}
To separate the effectiveness of the interaction strategy from the capability
of a particular video generator, we evaluate each method with three video-generation backbones:
\texttt{Omni-1.1-Flash}~\citep{google2026omni11}, \texttt{Veo~3.1}~\citep{deepmind2025veo31}, and a locally hosted
\texttt{MiniMax-H3}~\citep{minimax2026h3}.

\subsection{Evaluation Results}
\label{sec:results}

Table~\ref{tab:main_results} summarizes the main evaluation results across
three video-generation backbones. Overall, \MethodName{} improves contextual grounding
while maintaining strong visual quality, achieving the highest composite
score with every backbone.


\definecolor{headerblue}{RGB}{31,48,108}
\definecolor{heatblue}{RGB}{109,177,193}

\newcolumntype{L}[1]{>{\raggedright\arraybackslash}m{#1}}
\newcolumntype{C}[1]{>{\centering\arraybackslash}m{#1}}

\newcommand{\best}[1]{\cellcolor{heatblue!36}\textbf{#1}}
\newcommand{\second}[1]{\cellcolor{heatblue!28}\underline{#1}}
\newcommand{\third}[1]{\cellcolor{heatblue!19}#1}
\newcommand{\fourth}[1]{\cellcolor{heatblue!11}#1}
\newcommand{\fifth}[1]{\cellcolor{heatblue!4}#1}

\newcommand{\oursname}{%
    \textcolor{headerblue}{\textbf{\MethodName}}%
}

\newcommand{\compci}[3]{#1\, {[#2, #3]}}

\begin{table}[t]
\centering
\scriptsize
\setlength{\tabcolsep}{1.4pt}
\renewcommand{\arraystretch}{1.16}

\begin{adjustbox}{width=\textwidth,center}
\begin{tabular}{
    @{}
    L{2.55cm}
    *{6}{C{1.08cm}}
    @{\hspace{5pt}}
    *{6}{C{1.08cm}}
    @{\hspace{5pt}}
    C{2.65cm}
    @{}
}
\toprule
&
\multicolumn{6}{c}{\textcolor{headerblue}{\textbf{Narrative Grounding}}}
&
\multicolumn{6}{c}{\textcolor{headerblue}{\textbf{Knowledge Grounding}}}
&
\multicolumn{1}{c}{\textcolor{headerblue}{\textbf{Overall}}}
\\[-1pt]

\arrayrulecolor{headerblue!50}
\cmidrule(lr){2-7}
\cmidrule(lr){8-13}
\cmidrule(l){14-14}
\arrayrulecolor{black}

&
\textbf{SF} & \textbf{SC} & \textbf{IF}
& \textbf{JCS} & \textbf{VC} & \textbf{VQ}
&
\textbf{SF} & \textbf{SC} & \textbf{IF}
& \textbf{JCS} & \textbf{VC} & \textbf{VQ}
&
\textbf{Comp. [CI]}
\\ [1pt]

\midrule
\multicolumn{14}{c}{\textcolor{headerblue}{Backbone: \textbf{Omni-1.1-Flash}}}
\\

Direct Generation
& \fifth{83.9} & \third{66.1} & \fifth{80.4}
& \fifth{76.8} & \third{45.1} & \fifth{56.2}
& \fourth{59.4} & \second{68.8} & \fourth{62.5}
& \third{63.5} & \fourth{50.2} & \fifth{52.6}
& \third{\compci{70.2}{62.50}{77.83}}
\\

MM-StoryAgent
& \second{90.6} & \fourth{59.4} & \fourth{81.2}
& \fourth{77.1} & \fourth{44.0} & \third{59.4}
& \fifth{37.5} & \fifth{34.4} & \fifth{40.6}
& \fifth{37.5} & \second{53.4} & \second{64.9}
& \fifth{\compci{57.3}{41.15}{73.44}}
\\

MovieAgent
& \fourth{85.4} & \second{67.7} & \third{85.4}
& \second{79.5} & \second{46.6} & \second{60.9}
& \third{65.6} & \fourth{46.9} & \third{65.6}
& \fourth{59.4} & \fifth{47.1} & \fourth{60.4}
& \fourth{\compci{69.4}{57.29}{81.60}}
\\

VideoGen-of-Thought
& \third{89.6} & \fifth{56.3} & \second{86.5}
& \third{77.4} & \fifth{42.7} & \fourth{58.1}
& \second{69.5} & \third{54.2} & \best{75.3}
& \second{66.3} & \third{52.4} & \third{63.4}
& \second{\compci{71.9}{64.50}{78.70}}
\\

\oursname
& \best{92.7} & \best{81.3} & \best{90.6}
& \best{88.2} & \best{53.0} & \best{62.4}
& \best{87.5} & \best{78.1} & \second{75.0}
& \best{80.2} & \best{65.4} & \best{68.2}
& \best{\compci{84.2}{77.78}{90.45}}
\\[2pt]

\midrule
\multicolumn{14}{c}{\textcolor{headerblue}{Backbone: \textbf{Veo 3.1}}}
\\

Direct Generation
& \fourth{76.8} & \second{62.5} & \fourth{80.4}
& \third{73.2} & \third{39.2} & \fifth{49.7}
& \fourth{62.5} & \second{70.8} & \third{62.5}
& \second{65.3} & \third{49.7} & \fifth{58.5}
& \second{\compci{69.3}{57.34}{79.37}}
\\

MM-StoryAgent
& \fifth{70.5} & \fifth{41.1} & \fifth{70.5}
& \fifth{60.7} & \fifth{32.2} & \fourth{56.1}
& \fifth{50.0} & \third{66.7} & \fourth{60.4}
& \fifth{59.0} & \best{57.4} & \fourth{62.4}
& \fifth{\compci{59.9}{45.73}{72.07}}
\\

MovieAgent
& \second{89.3} & \fourth{57.2} & \best{86.6}
& \second{77.7} & \fourth{38.5} & \third{59.7}
& \third{64.6} & \fourth{56.3} & \fifth{58.3}
& \fourth{59.7} & \fourth{47.5} & \best{65.7}
& \third{\compci{68.7}{56.85}{76.39}}
\\

VideoGen-of-Thought
& \third{78.6} & \third{58.0} & \third{81.3}
& \fourth{72.6} & \second{43.7} & \second{60.1}
& \second{84.4} & \fifth{25.0} & \second{72.9}
& \third{60.8} & \fifth{44.3} & \third{62.5}
& \fourth{\compci{66.7}{58.55}{74.08}}
\\

\oursname
& \best{92.9} & \best{83.0} & \second{84.8}
& \best{86.9} & \best{55.9} & \best{62.4}
& \best{93.8} & \best{75.0} & \best{78.1}
& \best{82.3} & \second{54.8} & \second{64.2}
& \best{\compci{84.6}{80.95}{88.39}}
\\[2pt]

\midrule
\multicolumn{14}{c}{\textcolor{headerblue}{Backbone: \textbf{MiniMax-H3}}}
\\

Direct Generation
& \fifth{67.0} & \fifth{42.0} & \fifth{57.1}
& \fifth{55.4} & \fifth{34.3} & \third{45.2}
& \fifth{25.0} & \fifth{6.3} & \fifth{31.3}
& \fifth{20.8} & \fifth{20.5} & \fourth{45.8}
& \fifth{\compci{38.1}{27.23}{49.11}}
\\

MM-StoryAgent
& \second{78.6} & \third{57.1} & \best{80.4}
& \third{72.0} & \second{39.5} & \fourth{44.0}
& \third{81.3} & \fourth{31.3} & \fourth{65.6}
& \fourth{59.4} & \fourth{43.8} & \fifth{45.0}
& \third{\compci{65.7}{55.06}{76.34}}
\\

MovieAgent
& \second{78.6} & \second{65.2} & \third{75.0}
& \second{72.9} & \third{39.0} & \second{46.4}
& \best{93.8} & \second{58.3} & \best{81.3}
& \second{77.8} & \best{54.2} & \second{57.3}
& \second{\compci{75.4}{68.21}{81.70}}
\\

VideoGen-of-Thought
& \fourth{67.7} & \fourth{56.3} & \fourth{74.0}
& \fourth{66.0} & \fourth{37.9} & \fifth{42.6}
& \fourth{79.2} & \third{39.6} & \third{72.9}
& \third{63.9} & \third{45.5} & \third{50.2}
& \fourth{\compci{64.9}{57.64}{71.35}}
\\

\oursname
& \best{84.4} & \best{70.8} & \second{77.1}
& \best{77.4} & \best{46.1} & \best{47.8}
& \second{87.5} & \best{71.9} & \best{81.3}
& \best{80.2} & \second{46.7} & \best{58.7}
& \best{\compci{78.8}{75.52}{81.42}}
\\

\bottomrule
\end{tabular}
\end{adjustbox}

\caption{
\textbf{Main results.}
We compare each method under narrative and knowledge grounding across three video-generation backbones. Metrics include Source Faithfulness (SF), Story Continuity (SC), Interaction Fulfillment (IF), Joint Constraint Satisfaction (JCS), Visual Consistency (VC), and Video Quality (VQ).
\textbf{Comp.} denotes the overall composite score, followed by its confidence interval in brackets.
Darker cells indicate stronger performance within each backbone and metric.
\textbf{Bold} denotes the best result, while
\underline{underlining} denotes the second-best result.
Higher is better for all metrics.
}
\label{tab:main_results}
\end{table}

\begin{figure}[t]
    \centering
    \includegraphics[width=\linewidth]{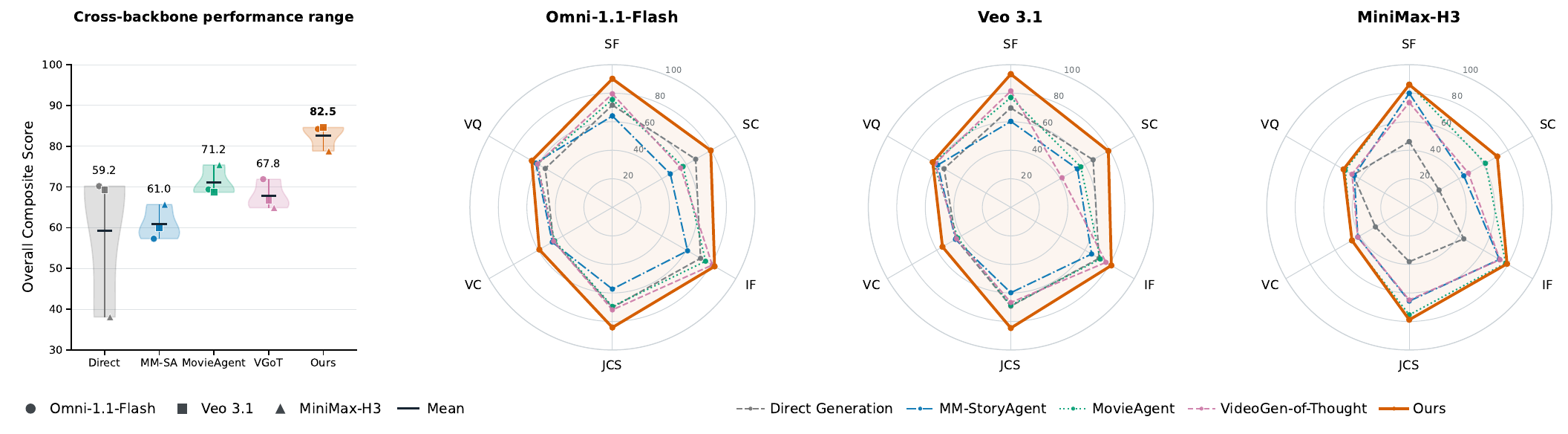}
    \caption{
    \textbf{Cross-backbone comparison.}
    \textbf{Left:} overall composite-score variation.
    \textbf{Right:} performance across the six evaluation dimensions,
    averaged over settings for each
    backbone.
    }
    \label{fig:cross_backbone}

\end{figure}

\begin{figure}[!t]
    \centering
    \includegraphics[width=\linewidth]{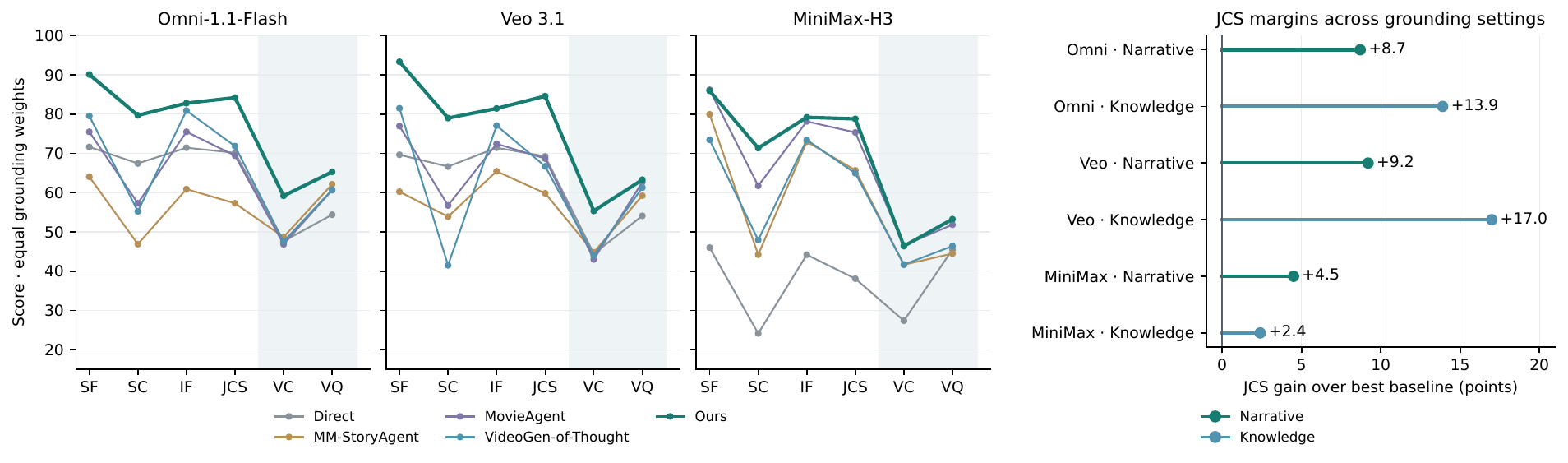}
\caption{\textbf{\MethodName{} Performance gain over the baselines.} Left: performance across six metrics for three video generators under equal grounding weights. Right: our method’s JCS gain over the strongest baseline in narrative and knowledge settings (percentage points).}
    \label{fig:compairsion}

\end{figure}

\noindent\textbf{Contextual grounding improves interactive continuation quality.}
The clearest gains appear on metrics that require the model to reconcile the
grounding source, previously rendered video, and current viewer request. Across
the grounding-related dimensions (SF, SC, IF, and \paperchange{JCS}), \MethodName{} improves
over the strongest competing result by roughly 10\% on average across
backbones and settings.


\noindent\textbf{The mechanism generalizes across grounding sources.}
We observe consistent improvements in both narrative- and knowledge-grounded
settings, even though the two settings emphasize different forms of
information. Narrative grounding requires maintaining characters, events, and
story progression, while knowledge grounding places greater emphasis on
accurately incorporating external information. In both cases, our method
retains a similar advantage over competing approaches.


\Needspace{7\baselineskip}
\noindent\textbf{The improvements are robust across video backbones.}
Figure~\ref{fig:cross_backbone} further shows that baseline performance can
change substantially with the underlying video generator. The composite score
of Direct Generation spans more than 30 points across the three backbones,
whereas the range of \MethodName{} is below 6 points. This corresponds to an
approximately 82\% reduction in cross-backbone performance variation relative to Direct Generation.



\FloatBarrier
\section{Discussion}
\label{sec:discussion}

\subsection{Case Study}
\label{sec:case-study}

\begin{figure}[!ht]
    \centering
    \includegraphics[width=\linewidth]{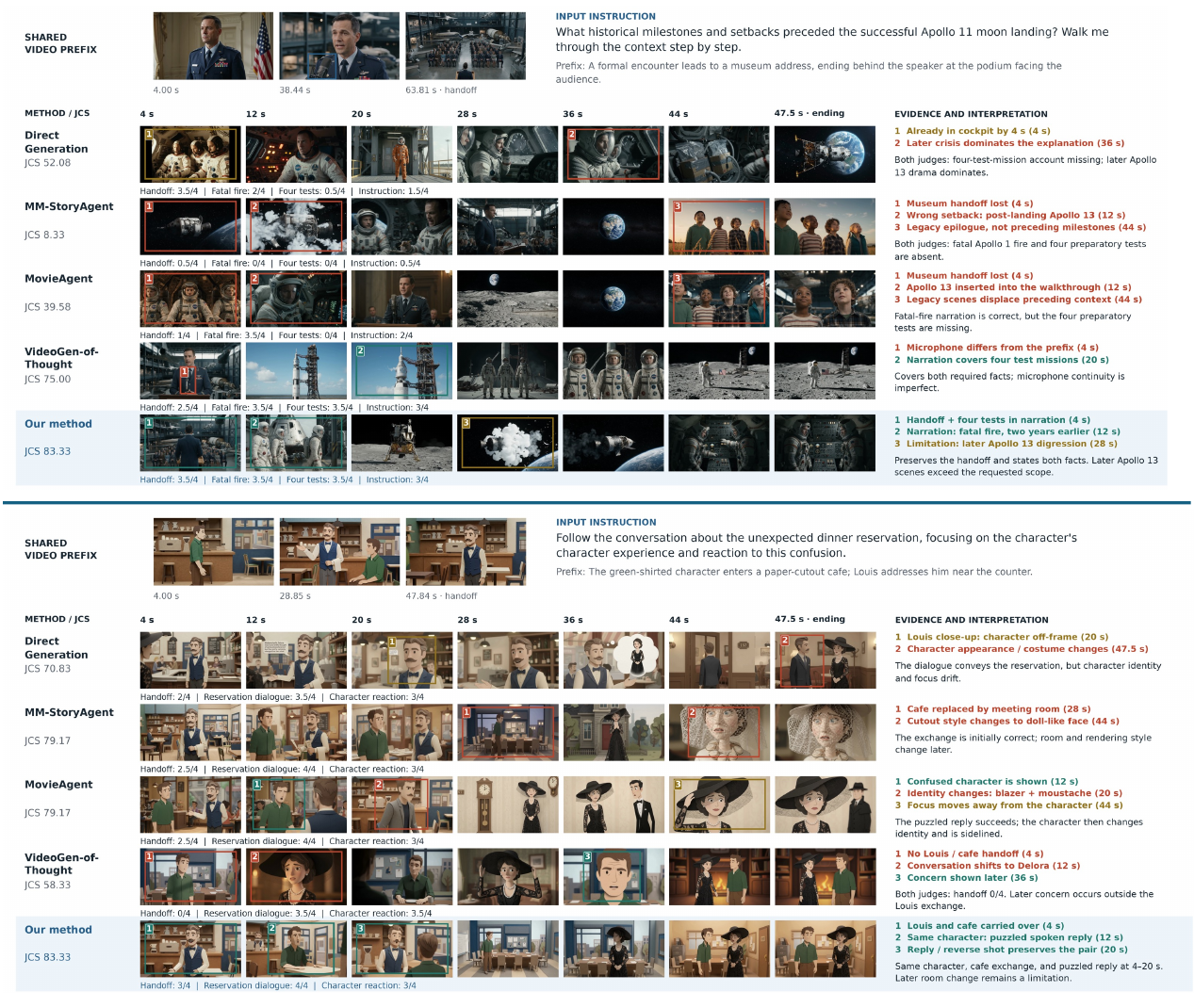}
    \caption{\textbf{Qualitative Example.}
    Colored boxes highlight successful evidence preservation and representative
    failures.}
    \label{fig:example}
\end{figure}

Figure~\ref{fig:example} illustrates how different methods balance the three sources of information in interactive storytelling. The example requires a continuation of an Apollo-related sequence while explaining the historical milestones and setbacks preceding the Apollo~11 landing. Several baselines capture individual facts but lose information already established by the prefix or introduce events in an inappropriate order. For example, MM-StoryAgent~\citep{xu2025mmstoryagent} quickly loses the museum handoff and omits much of the requested historical context, while MovieAgent~\citep{wu2025movieagent} introduces Apollo~13 before completing the requested preceding milestones. VideoGen-of-Thought provides substantially better factual coverage, but still exhibits visual discontinuity from the prefix. In contrast, our method preserves the handoff, introduces the Apollo~1 fire and preceding test missions in the requested narrative, and maintains a more coherent progression across the continuation.

\vspace{-0.2in}

\subsection{Ablation Study}
\label{sec:ablation}

\leavevmode{We examine two design choices in \MethodName{}. \emph{Unstructured State} replaces the structured representation $\mathcal{M}=(\mathcal{V},\mathcal{H},\mathcal{S},\mathcal{A})$ with a free-form summary while retaining the remaining pipeline. \emph{One-Pass Planning} plans directly from the structured representation without instance-specific constraint induction. In the narrative-grounded setting, One-Pass Planning and Unstructured State reduce the mean of SF, SC, and IF by 4.36 and 7.60 percentage points, respectively, relative to the full method. The observed reductions in narrative visual consistency for Unstructured State and knowledge-grounded interaction fulfillment for One-Pass Planning are consistent with complementary roles for context representation and dependency-guided planning. These comparisons assess the implemented configurations; they do not independently isolate representation, repair, and reasoning-budget effects. \appref{app:comparison-scope} explains their interpretation.}

\Needspace{24\baselineskip}
\subsection{Limitation and Failure Analysis}
\label{sec:failure-analysis}

\begin{wrapfigure}{l}{0.40\textwidth}
\centering
\includegraphics[width=\linewidth]{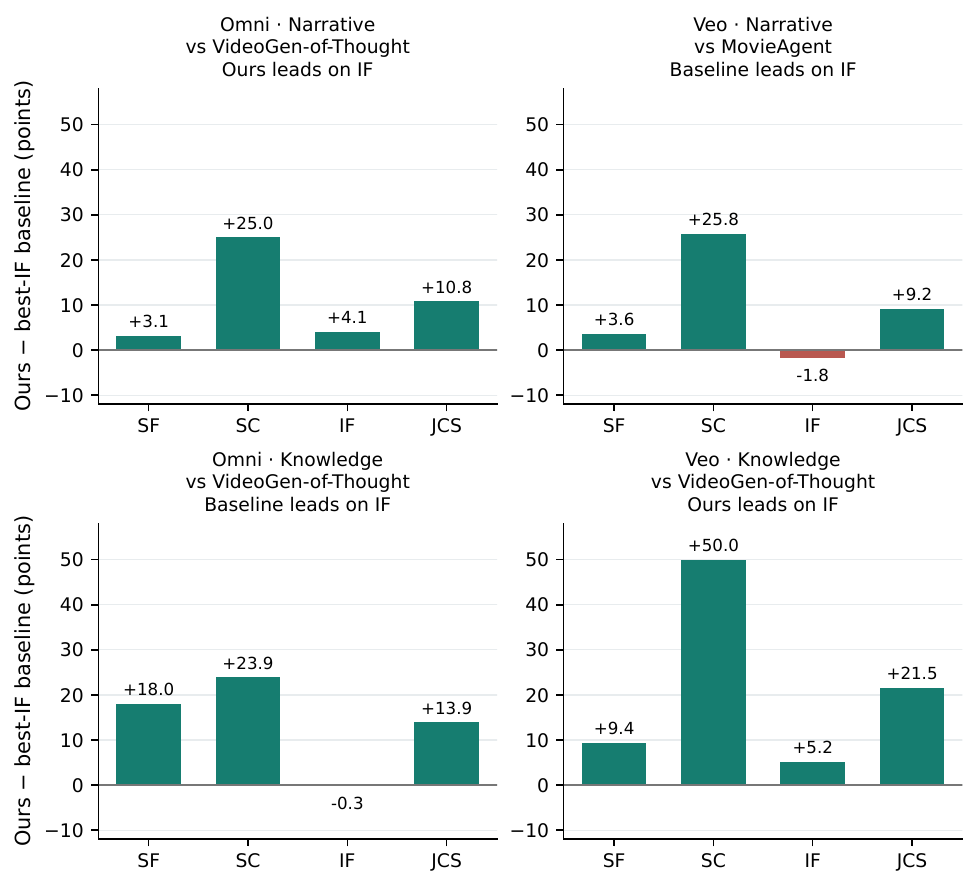}
\caption{The comparator is the IF-leading baseline, held fixed across all four metrics in each panel. }
\label{fig:limitation}
\end{wrapfigure}

\MethodName{} is not best on every individual metric, including some IF, VC, and VQ comparisons. For IF, our method jointly considers the viewer request with the grounding source and previously rendered video. When a request is ambiguous, aggressive, or partially conflicts with established context, the framework may favor a continuation that remains compatible with prior evidence instead of following the request as literally as possible. In contrast, methods that condition more directly on the latest instruction may obtain higher IF in such cases, but can sacrifice source faithfulness or story continuity. Similarly, visual quality remains largely bounded by the underlying video
generator because our method does not modify the synthesis model itself. These metric-level gaps therefore reflect the multi-objective nature of video interactive storytelling: maximizing a single criterion independently can be easier than satisfying the source, prior video, and viewer request jointly.
Despite these trade-offs, \MethodName{} achieves the strongest and most balanced
overall composite performance across backbones.

\Needspace{18\baselineskip}
\begin{wrapfigure}{r}{0.50\textwidth}
\centering

\includegraphics[width=\linewidth]{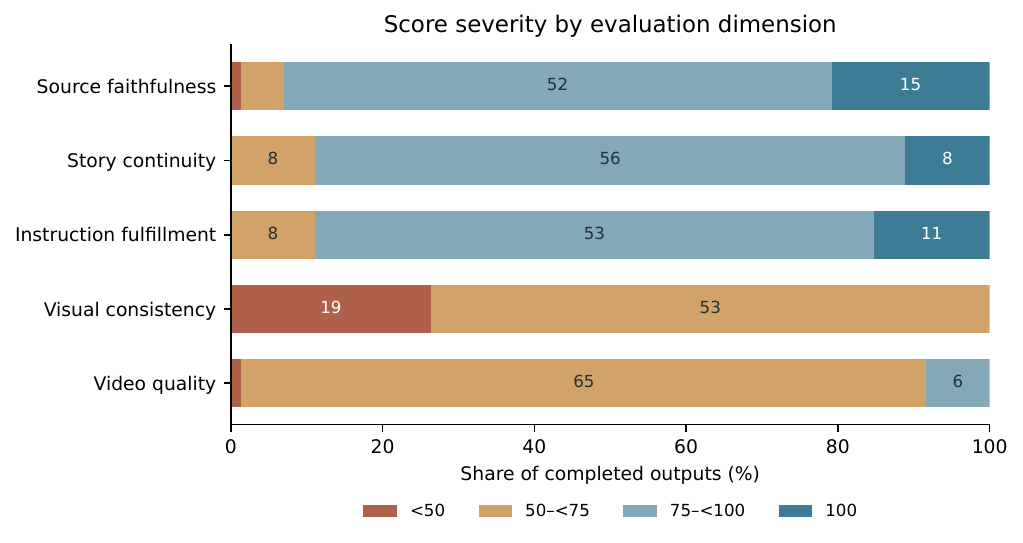}

\caption{{Score distribution in the 72-output analysis sample.}}
\label{fig:failure_analysis}

\end{wrapfigure}

\leavevmode{Figure~\ref{fig:failure_analysis} analyzes a sample of 72 completed outputs selected for failure analysis; this is an analysis subset, not the total benchmark evaluation set. Within this sample, 67 outputs score at least 75 on Source Faithfulness, while 64 reach that threshold on each of Story Continuity and Interaction Fulfillment.} Perfect scores are less common, however, especially for Story Continuity (8 outputs) and Interaction Fulfillment (11 outputs). {In this analysis sample, continuations often preserve the main context while missing details of the established sequence or request.}

\section{Conclusion}

We introduced video interactive storytelling with contextual grounding, along with \MethodName{} and \BenchName{}. By abstracting the source, video prefix, and viewer request into aligned states and inducing constraints for generation, \MethodName{} improves performance across three video backbones.

\clearpage
\bibliography{main}
\clearpage
\appendix
\section{Related Work}
\label{app:related-work}
\label{sec:related-work}

\noindent\textbf{Video Storytelling and Creative Collaboration.}
Video storytelling systems coordinate narrative planning and multimodal
generation to produce coherent visual stories.
MM-StoryAgent~\citep{xu2025mmstoryagent} orchestrates text, images, and
audio for narrated storybook videos, while
MovieAgent~\citep{wu2025movieagent} generates multi-scene movies from
scripts and character references.
VideoGen-of-Thought~\citep{zheng2024videogenofthought},
Co-Director~\citep{song2026codirector}, and
ViMax~\citep{vimax2026} further explore structured planning, agent
coordination, and refinement for coherent video production.
These approaches primarily organize generation around an initial
creative brief, script, or narrative source.
Complementary human--AI tools support interaction during creative production:
CoLyricist~\citep{yoshida2026colyricist} provides workflow-aligned assistance
for theme setting, ideation, drafting, and melody fitting.
Our setting instead considers requests made after part of the video
has been generated: each continuation must accommodate the new request
while remaining faithful to the grounding source and consistent with
the rendered video prefix.

\noindent\textbf{Interactive Video Generation and World Models.}
Streaming video models enable viewers to influence generation as it unfolds. CausVid~\citep{yin2025causvid}, StreamDiT~\citep{kodaira2025streamdit}, and LongLive~\citep{yang2025longlive} support evolving text prompts, while StreamDiffusionV2~\citep{streamdiffusionv2} supports dynamic generation with streaming inputs. Interactive world models extend this responsiveness to action-conditioned environments: Genie~\citep{bruce2024genie}, Oasis~\citep{oasis2024}, Matrix-Game~2.0~\citep{matrixgame2025}, and Vid2World~\citep{huang2025vid2world0} generate worlds that evolve with viewer actions, and Genie~3~\citep{deepmind2025genie3} also supports promptable world events. Together, these systems advance responsive generation and continuity across interactions. Video interactive storytelling additionally requires interpreting each new request in light of both an external grounding source and the story already rendered, then using their joint context to plan the continuation.

Long-horizon multimodal agents likewise need to retain information relevant
to evolving decisions. MeMento~\citep{li2026memento} compresses multimodal
history according to user preferences for embodied decision-making.
Our structured context serves a different role: it preserves grounded states,
evidence links, and cross-context dependencies from which constraints on a
candidate video continuation can be induced.

\noindent\textbf{Constrained Multimodal Generation and Repair.}
Structured controls provide an interface between user intent and generation.
Divide and Conquer Generation~\citep{li2024divideconquer} addresses lexical
constraint satisfaction by separating satisfied and missed constraints and
combining generated responses. REFFLY~\citep{zhao2025reffly} revises lyrics
to fit a melody while preserving semantic meaning.
EchoFoley~\citep{li2026echofoley} represents sounding events symbolically
to support hierarchical control over video-grounded audio generation.
These approaches address lexical, musical, or acoustic controls.
\MethodName{} instead derives constraints from dependencies that emerge
jointly from the grounding source, rendered video prefix, and viewer request.

Agentic refinement complements such control mechanisms.
METAL~\citep{li2025metal} coordinates specialized agents to generate,
critique, and revise charts under a computational budget.
VISCO~\citep{wu2025visco} evaluates fine-grained visual reasoning critique
and correction, showing that model-generated critiques can themselves be
unreliable. AgentDebug~\citep{zhu2025agentdebug} identifies root-cause
failures in agent trajectories and provides targeted corrective feedback.
Our repair loop connects these concerns to video continuation by tracing
script and rendered-output violations to the same grounded dependencies.

\Needspace{7\baselineskip}
\noindent\textbf{Video Generation and Storytelling Benchmarks.}
Existing benchmarks evaluate complementary aspects of visual
generation.
VBench~\citep{huang2024vbench} measures video generation quality;
StoryBench~\citep{bugliarello2023storybench} evaluates action execution,
story continuation, and story generation;
and MovieBench~\citep{wu2025moviebench} provides hierarchical movie
data for studying long-form generation.
ViStoryBench~\citep{zhuang2025vistorybench} evaluates narrative
alignment and consistency across image sequences, while
GenAD-Bench~\citep{song2026codirector} evaluates generative advertising
workflows.
Tailor-Bench~\citep{li2026tailorbench} probes irregular physical interactions
through regular, unconventional, and impossible scenarios, revealing
limitations in state-change realization and temporal consistency.
More closely related, WBench~\citep{ying2026wbench} evaluates
interactive world models across setting adherence, interaction
adherence, consistency, physics compliance, and video quality,
covering navigation, subject actions, event editing, and perspective
switching.
Our benchmark focuses on the joint satisfaction of source
faithfulness, story continuity, and interaction fulfillment in
video continuations grounded in narrative or knowledge sources.

\begin{table}[htbp]
\centering
\footnotesize
\setlength{\tabcolsep}{4pt}
\renewcommand{\arraystretch}{1.1}

\begin{tabularx}{\linewidth}{@{}Xlccc@{}}
\toprule
& \textbf{Interaction Type}
& \makecell{\textbf{Source}\\\textbf{Faithfulness}}
& \makecell{\textbf{Story}\\\textbf{Continuity}}
& \makecell{\textbf{Interaction}\\\textbf{Fulfillment}} \\
\midrule

\multicolumn{5}{@{}l}{
    \color{black}\textit{Video storytelling methods}
} \\
\addlinespace[2pt]

MM-StoryAgent~\citep{xu2025mmstoryagent}
& ---
& --- & $\checkmark$ & --- \\

MovieAgent~\citep{wu2025movieagent}
& ---
& $\checkmark$ & $\checkmark$ & --- \\

VideoGen-of-Thought~\citep{zheng2024videogenofthought}
& ---
& --- & $\checkmark$ & --- \\

Co-Director~\citep{song2026codirector}
& ---
& $\triangle^{a}$ & $\checkmark$ & --- \\

ViMax~\citep{vimax2026}
& ---
& $\checkmark^{b}$ & $\checkmark$ & --- \\


CausVid~\citep{yin2025causvid}
& Free-Form
& --- & $\triangle^{d}$ & $\checkmark$ \\

StreamDiT~\citep{kodaira2025streamdit}
& Free-Form
& --- & $\triangle^{d}$ & $\checkmark$ \\

LongLive~\citep{yang2025longlive}
& Free-Form
& --- & $\triangle^{d}$ & $\checkmark$ \\

StreamDiffusionV2~\citep{streamdiffusionv2}
& Free-Form
& --- & $\triangle^{d}$ & $\triangle^{f}$ \\

Genie~\citep{bruce2024genie}
& Limited Action Space
& --- & --- & $\triangle^{c}$ \\

Oasis~\citep{oasis2024}
& Limited Action Space
& --- & --- & $\triangle^{c}$ \\

Matrix-Game 2.0~\citep{matrixgame2025}
& Limited Action Space
& --- & --- & $\triangle^{c}$ \\

Vid2World~\citep{huang2025vid2world0}
& Limited Action Space
& --- & --- & $\triangle^{c}$ \\


\midrule
\multicolumn{5}{@{}l}{
    \color{black}\textit{Video generation and visual storytelling benchmarks}
} \\
\addlinespace[2pt]

MovieBench~\citep{wu2025moviebench}
& ---
& $\triangle^{a}$ & $\triangle^{d}$ & --- \\

ViStoryBench~\citep{zhuang2025vistorybench}
& ---
& $\triangle^{a}$ & $\triangle^{d,g}$ & --- \\

GenAD-Bench~\citep{song2026codirector}
& ---
& $\triangle^{a}$ & --- & --- \\

VBench~\citep{huang2024vbench}
& ---
& --- & --- & --- \\

WBench~\citep{ying2026wbench}
& Free-Form
& $\triangle^{e}$ & --- & $\checkmark$ \\

StoryBench~\citep{bugliarello2023storybench}
& Free-Form
& --- & $\triangle^{d}$ & $\triangle^{c}$ \\

\midrule

\textbf{Video Interactive Storytelling with Contextual Grounding}
& Free-Form
& $\checkmark$ & $\checkmark$ & $\checkmark$ \\

\bottomrule
\end{tabularx}
\caption{\textbf{Comparison with related methods and benchmarks.}
$\checkmark$: explicitly addressed;
$\triangle$: partial coverage;
---: not explicitly established.
$^{a}$only asset or script alignment;
$^{b}$only screenplay adaptation;
$^{c}$action-conditioned control;
$^{d}$only temporal coherence;
$^{e}$only world-setting adherence;
$^{f}$live video interaction;
$^{g}$image-sequence evaluation.}
\label{tab:related-work}

\end{table}

\FloatBarrier

\section{Benchmark Construction and Data Interface}
\label{app:benchmark}

This appendix describes benchmark construction, grounded continuation, evaluation, and study materials for video interactive storytelling. The continuation setting uses two 24-second scenes, each comprising three eight-second shots. Illustrative examples are distinguished from empirical results.


\subsection{Source collection and instance counts}
\label{app:data-counts}

\BenchName{} is constructed from 250 CoQA~\citep{coqa} source passages drawn from MCTest, Gutenberg, RACE, CNN, and Wikipedia. The collection contains 132 narrative-grounded passages and 118 knowledge-grounded passages. Narrative grounding preserves the characters, events, and relations specified by a core story; knowledge grounding preserves factual claims and their qualifications in an external information source. Each passage is converted into a scene-based story with aligned source facts and rendered in cinematic realism and cutout animation.

The construction comprises 1,004 interaction instructions at designated interruption points. Pairing each instruction with both visual styles gives 2,008 \emph{planned} instruction--style instances. These counts describe distinct units: a passage can support several instructions, and an instruction can appear in two styles.

Because each configuration requires rendering a 48-second multi-shot continuation, we evaluate a fixed subset of \BenchName{} rather than the full 2,008 instances. For each of the three backbones, we sample 8 narrative-grounded and 8 knowledge-grounded instances from distinct source passages, and run all five methods on the same 16 instances, yielding 240 evaluated continuations in total ($5\ \text{methods} \times 3\ \text{backbones} \times 16\ \text{instances}$).

\subsection{Evaluation setting and inputs}
\label{app:data-interface}

All benchmark instances belong to the evaluation split, no training split is used.

An interactive continuation instance consists of a grounding source $\mathcal{G}$, an observed video prefix $v_{\mathrm{prefix}}$, and the current viewer instruction $u$, together with relevant reference assets. The source remains persistent context, while the prefix records the visual and narrative state already shown to the viewer. The continuation must respond to $u$ without contradicting either source-supported content or established history. The task formulation does not separately condition on the complete history of earlier viewer instructions; it relies on the rendered prefix to carry the relevant observable history.

Target facts and continuation requirements are reserved for evaluation. This distinction is essential: the generator may derive its own context-dependent constraints from its permitted inputs, but access to the evaluator's answer annotations or reference checklist would change the task.

\section{Grounded Dependencies and Continuation Repair}
\label{app:method}

\subsection{Representation and evidence provenance}
\label{app:representation}

The purpose of context compilation is to expose dependencies that cannot be recovered reliably from the latest instruction alone. We retain the main paper's notation $\mathcal{M}=(\mathcal{V},\mathcal{H},\mathcal{S},\mathcal{A})$ for aligned contextual states: $\mathcal{V}$ describes the observed video, $\mathcal{H}$ the evolving story, $\mathcal{S}$ source-supported content, and $\mathcal{A}$ cross-context alignments. \paperchange{We derive the dependency set $\mathcal{R}=\operatorname{Derive}(\mathcal{M})$ from these aligned states. Thus, $\mathcal{R}$ is a derived object rather than an additional component of $\mathcal{M}$, separating the evidence representation from relations inferred over it.}

A grounded predicate identifies an entity or event, its attributes or relations, its temporal scope, and supporting evidence. Entity alignment associates references to the same character, object, event, or claim across modalities. Temporal scope is necessary because an earlier observation may no longer describe the current state: possession can change after a transfer, and an action can alter an object's location or condition. Source predicates also retain qualifications and conditions so that a restricted claim is not promoted to an unconditional fact.

Evidence provenance provides the connection between a verification decision and the context that justifies it. Source evidence can identify a supporting passage; video evidence can identify the observation supporting a state or transition. An inferred dependency should remain distinguishable from a directly observed fact. The implementation schemas, prompts, and agent/planner code map this interface to source facts, a prefix ledger, source alignment, instruction requirements, a scene script, and an expanded shot plan. Source alignment separates shown, partially shown, unseen, and uncertain facts. The shot plan records prompts, reference identifiers, audio, camera, and transitions; review traces record defects, revised plans, and selected attempts. The conceptual state model is thus represented through several structured intermediate artifacts.



\subsection{Candidate-conditioned activation}
\label{app:activation}

For a candidate script $\mathcal{P}^{(k)}$, a dependency is projected into a constraint $c_j=(s_j,a_j,b_j,e_j)$ with scope $s_j$, applicability condition $a_j$, required relation $b_j$, and supporting evidence $e_j$. The active constraint set is
\begin{equation}
\mathcal{C}^{(k)}=\left\{c_j:\;a_j(\mathcal{P}^{(k)},\mathcal{M})=1\right\}.
\end{equation}
This notation does not prescribe a new activation classifier; it makes explicit the candidate-conditioned operation described in the method. A dependency that is irrelevant to the proposed events need not constrain the continuation. When a repair introduces a new event, activation must be recomputed because the new event can introduce additional prerequisites.

Each active constraint receives one of three outcomes: \texttt{pass}, \texttt{fail}, or \texttt{unknown}. A failure indicates an identifiable conflict or missing prerequisite; an unknown outcome indicates insufficient evidence to decide compatibility. These outcomes must remain distinct. In particular, a lack of evidence should not be reported as verified satisfaction. The main method includes failed and unresolved dependencies in the feedback used to revise a script.

\subsection{Illustrative dependency trace}
\label{app:trace}

Consider a hypothetical story in which a cabinet can only be opened with a key. The rendered prefix establishes that character $A$ has transferred the key to character $B$, and the viewer asks to see $A$ open the cabinet. This example illustrates the mechanism; it is not presented as a sampled benchmark instance or an empirical result.

The source supplies the opening prerequisite, while the prefix supplies the latest possession state. A candidate in which $A$ immediately uses the key activates a dependency that conflicts with the established transfer. A possible repair is to show $B$ returning the key before $A$ opens the cabinet, provided this event is compatible with the source and story. Verification must then check the revised event sequence, rather than merely check whether the word ``key'' appears in the script.

Rendering introduces a separate point of failure. A repaired script may include the return of the key even when the synthesized video omits that action. Output verification therefore checks the realized relations against the same grounded dependencies. If the transfer is not visually established, the output does not demonstrate the planned prerequisite, and the feedback can identify the missing transition for a subsequent repair.

\Needspace{6\baselineskip}
\subsection{Repair loop and termination}
\label{app:repair}

\paperchange{The overall procedure, including Constrained Video Generation, is summarized as follows:}
\begin{enumerate}[leftmargin=*,itemsep=2pt,topsep=3pt]
\item Extract and align video, story, and source predicates; derive contextual dependencies.
\item Propose a candidate continuation script conditioned on the grounded context and current request.
\item Activate candidate-relevant constraints and evaluate their compatibility with the script.
\item Use failed or unresolved dependencies and their evidence to produce targeted feedback; revise the script and recompute activation.
\item Render a script that satisfies the applicable script-level constraints, and inspect the realized video relations.
\item Repair missing or conflicting realizations and render again, stopping when verification succeeds or the computation budget is exhausted.
\end{enumerate}

This procedure is a control flow, not a proof of convergence or correctness. Extraction can miss evidence, inferred dependencies can be wrong, and a renderer can repeatedly fail to realize a valid script. Budget exhaustion must therefore be distinguished from verified success. Similarly, a request incompatible with persistent source constraints need not admit a continuation that fully satisfies all objectives.

\subsection{Two-scene implementation}
\label{app:runtime}

\paragraph{State extraction and structural repair.}
The base agent extracts ordered source facts with supporting quotations, inspects the complete prefix into a conservative ledger, partitions source facts into shown/partially shown/unseen/uncertain categories, and decomposes the request into source-linked requirements. Uncertain observations are kept separately, and unsupported requests are explicitly recorded. The script validator is called after the first proposal and each repair, with at most five repairs. The agent returns the validation result alongside its output, so reaching the repair limit does not itself establish validity.

\paragraph{Plan review.}
The six-shot refinement function performs at most three reviews and two repairs. It supplies the actual prefix video, public source, request, style, assets, reference limit, public states, and candidate plan to the reviewer. Acceptance requires an affirmative acceptance flag and empty source/instruction, continuity, and renderability error lists. Each repair receives those defects, after which asset validation is rerun. The function records the review history and final pass flag; after its last review it returns the plan even if that flag is false. Consequently, the orchestration must distinguish a budget-exhausted plan from an accepted plan.

\paragraph{Boundary and audio contracts.}
The first shot uses prefix continuation. Later shots use exact continuation or a motivated hard cut; the first shot of scene 2 follows the explicit inter-scene boundary. Shot normalization restricts fact and requirement references to the corresponding scene and restricts asset references to valid scene assets. A speaking scene assigns speech to exactly one of its three shots; a silent scene uses ambience only. The optional speech audit checks six ordered shots, requires verbatim source quotations for supported factual speech, and allows at most three audit calls before raising a failure.

\begin{table}[htbp]
\centering\small
\caption{Shared model configuration for continuation generation and evaluation.}
\label{tab:runtime-models}
\begin{tabular}{lp{0.64\linewidth}}
\toprule Role & Model identifier \\
\midrule
Planner & \texttt{gemini-3.1-pro-preview}~\citep{google2026gemini31}\\
Reference-image generator & \texttt{gemini-2.5-flash-image}~\citep{google2025flashimage}\\
Gemini evaluator & \texttt{gemini-3.8-flash}~\citep{google2026gemini38}\\
Claude evaluator & \texttt{claude-opus-5}~\citep{anthropic2026opus5}\\
Omni renderer & \texttt{gemini-omni-1.1-flash-preview}~\citep{google2026omni11}\\
Veo renderer & \texttt{veo-3.1-generate-001}~\citep{deepmind2025veo31}\\
MiniMax renderer & \texttt{MiniMax-H3-Ref2VA-Turbo-4step-v0.1-bf16}~\citep{minimax2026h3}\\
\bottomrule
\end{tabular}
\end{table}
The shared configuration caps agent calls at 48, image candidates at three, input tokens at 600,000, and output/reasoning tokens at 96,000. These are upper limits rather than measured mean usage.

\section{Evaluation Checklists and Score Interpretation}
\label{app:evaluation}

\subsection{{Internal quality control and benchmark separation}}
\label{app:internal-quality}
{Generation-time review and attempt selection use internal quality checks, not the benchmark's evaluation criteria or metric scores. Their role is to identify defects in a proposed plan or rendered attempt using the public source, observed prefix, viewer request, assets, and intermediate states. They do not use the benchmark's private answer annotations, reference checklists, or final evaluator ratings to choose or repair an output.

Internal acceptance flags and quality scores therefore belong to the generation procedure. Final benchmark evaluation is a separate assessment of the selected output under the fixed benchmark rubric described below. Both procedures can inspect related properties, such as continuity or whether a requested action is visible, because those properties are part of the task. This shared task relevance does not give the generator access to benchmark labels or make final benchmark scores its optimization objective. In particular, generation-time quality review must not be interpreted as selecting outputs with the benchmark evaluator.}

\subsection{Metric-specific evidence}
\label{app:rubric}

The semantic metrics assess different sources of obligation. Source Faithfulness (SF) uses source-specific checks primarily derived from the original QA annotations, including facts, events, relationships, and causal dependencies. Story Continuity (SC) concerns established events, identities, object states, spatial arrangements, and causal history in the prefix. Interaction Fulfillment (IF) decomposes the latest request into atomic requirements such as subjects, actions, concepts, explanations, and presentation requirements. Keeping these checklists separate makes it possible to identify a continuation that satisfies the request while losing source or prefix information.

\begin{table}[htbp]
\centering
\begin{tabular}{cl}
\toprule
Ordered level & Interpretation \\
\midrule
0 & Absent or contradictory \\
1 & Mostly unsuccessful \\
2 & Partially fulfilled \\
3 & Mostly fulfilled \\
4 & Fully fulfilled \\
\bottomrule
\end{tabular}
\caption{Semantic rating anchors. Unknown evidence is recorded separately.}
\label{tab:app-rating}
\end{table}

The same rubric is used for 24-second and 48-second continuations, with evidence timestamps interpreted over the corresponding full continuation. A common transcript supplies speech evidence; sampled frames supply visual evidence. Each rating should be accompanied by evidence for the corresponding check. Evidence of an intended action in a script and evidence of a realized action in a video answer different questions. The documented evaluation protocol requires evidence from the rendered output: a required visible action cannot be satisfied solely by narration or by inclusion in a plan. Fixed check identifiers, applicability, metric membership, and evidence modality are determined independently of method outputs.

Visual Consistency (VC) assesses recurring characters, objects, environments, and style, both across the prefix--continuation boundary and within the continuation. Video Quality (VQ) assesses visible image integrity and composition in the sampled frames. These frame-based scores do not establish fine-motion or audio quality. These perceptual criteria should not substitute for semantic checks: a visually polished continuation can still contradict its source or omit the requested explanation.

\subsection{Normalization and judge aggregation}
\label{app:aggregation}

The documented scoring protocol codes each applicable semantic check as $r\in\{0,1,2,3,4\}$ and normalizes complete ratings as
\begin{equation}
M_i^{(q)}=\frac{25}{K_{i,M}}\sum_{k=1}^{K_{i,M}}r_{i,k}^{(q)},
\qquad M_i=\frac{M_i^{(1)}+M_i^{(2)}}{2}.
\end{equation}
Here $K_{i,M}>0$ is the number of fixed applicable checks for metric $M$ on instance $i$. Applicability is not changed after observing a method output. The two judges independently supply ratings and evidence, and their metric scores are averaged.

Unknown evidence is represented as null rather than a zero or midpoint. Incomplete applicable ratings produce a missing complete score; an observed-only score and coverage can be reported separately. If the known-rating sum is $R$ and $u$ of the $K$ applicable ratings remain unresolved, the documented lower and upper bounds are $25R/K$ and $25(R+4u)/K$. These bounds describe the contribution of unresolved checks.

\subsection{Joint score and overall composite}
\label{app:composite}

Joint Constraint Satisfaction is defined in the manuscript as
\begin{equation}
\mathrm{JCS}=\frac{\mathrm{SF}+\mathrm{SC}+\mathrm{IF}}{3}.
\end{equation}
Despite its name, this arithmetic mean is not the fraction of outputs that satisfy every constraint. A high score in one semantic dimension can compensate numerically for a low score in another. The individual semantic metrics must therefore remain visible alongside JCS.

The documented overall composite is the equal-weight mean of setting-level mean JCS:
\begin{equation}
C_{\mathrm{overall}}=\frac12\left(\frac{1}{n_N}\sum_{i\in N}\mathrm{JCS}_i+
\frac{1}{n_K}\sum_{i\in K}\mathrm{JCS}_i\right).
\end{equation}
VC and VQ are reported separately and do not contribute to this composite. Repeated styles or interventions derived from a single source are not independent source observations.

Although the planner shares a model family with one evaluator, the ranking of methods is unchanged when scored by Claude Opus~5 alone, the internal reviewer never sees the evaluation rubric or benchmark labels, and JCS is reported alongside a conjunctive variant (the fraction of instances with SF, SC, and IF all $\geq 75$) that yields the same ordering.

\section{Baseline Adaptations and Implementation Reporting}
\label{app:baselines}

The four baselines cover direct generation, coordinated story production, hierarchical planning, and intermediate reasoning. Their original formulations do not by themselves specify how to resume a partially rendered story after a viewer intervention. An interactive adaptation must therefore state where the source, observed prefix, and new request enter each pipeline and how the continuation is attached to the prefix.

\paragraph{Direct Generation.}
Mechanical shot prompts insert the public context, requested action, and temporal interval without an intermediate LLM story-planning call. The method still uses rendering and generation-time QA.

\paragraph{MM-StoryAgent.} \citep{xu2025mmstoryagent}
Up to three writer--expert exchanges precede narrative outlining and chapter expansion. Image, sound, speech, and music roles feed a shared shot plan and keyframe design. The expert observes the original prefix video. Audio-role text is realized through the common renderer's native audio, rather than separate upstream synthesis pipelines.

\paragraph{MovieAgent.} \citep{wu2025movieagent}
Source-to-script preparation observes the original prefix and is followed by screenwriting, scene planning, shot planning, and keyframe design. Public references replace a separate character bank; upstream lip synchronization is not reproduced. No additional storyline-repair pass follows these stages.

\paragraph{VideoGen-of-Thought.} \citep{zheng2024videogenofthought}
An initial outline observes the prefix, followed by character specifications and elaboration into Character, Background, Relation, Camera Pose, and HDR Description. Up to two critique/refinement rounds precede keyframe design. Native latent propagation, IP-adapter behavior, and smoothing are not reproduced through the common endpoint.

\subsection{{Comparison design and interpretation}}
\label{app:comparison-scope}
{\textbf{Why these baseline families.} Direct generation, multi-agent story production, hierarchical planning, and intermediate reasoning provide complementary ways to turn the same source, prefix, and request into a continuation. Adapting them tests whether these established planning strategies support contextual grounding under a shared video-generation interface. Interactive streaming methods such as StreamDiT and LongLive~\citep{kodaira2025streamdit,yang2025longlive} instead emphasize low-latency generation with evolving prompts. Their ability to generate long streams is not disputed; the comparison here concerns source-grounded, multi-shot continuation from a supplied prefix, rather than streaming throughput or maximum video duration.

\textbf{What the adaptations control.} The common rendering interface separates the planning strategy from the chosen synthesis backbone. The adaptations preserve the corresponding story-planning or reasoning organization while translating its output into the shared shot interface. Renderer-specific features that cannot be expressed through that interface, including latent propagation, separate lip synchronization, and smoothing, are explicitly identified above. The results consequently compare these adapted pipelines on the present task; they are not evaluations of the original systems with every native component intact.

\textbf{What the ablations establish.} Unstructured State probes the role of explicit contextual structure by replacing it with a free-form summary. One-Pass Planning probes the combined role of candidate-specific constraints and the associated planning procedure. These comparisons help interpret the implemented pipeline, but they do not separately estimate the causal effect of every representation field, repair stage, or reasoning call. In particular, the one-pass comparison should not be read as a compute-matched test of constraint induction alone.

\textbf{Computation and claim scope.} Shared inputs and video backbones control access to task information and the synthesis endpoint. They do not imply identical realized token usage, agent calls, retries, or latency across planning strategies. The runtime limits reported in \appref{app:runtime} specify configuration caps rather than measured equal expenditure. The reported comparisons therefore support conclusions about the evaluated end-to-end configurations, without establishing an efficiency advantage or attributing the entire gain to contextual representation independently of computation.}

\section{Formative Study and Human Evaluation}
\label{app:human}

\subsection{Formative study}
\label{app:formative}

{The formative study enrolled nine volunteers. Two participants did not pass the attention test and were excluded from the learning-gain analysis, leaving seven participants in the analyzed sample. Among these seven participants, four were master's students in computer science, biology, and physics, and three were AI scientists working in industry.} Learning gains and experience ratings measure distinct outcomes: the former assess changes in content understanding, while the latter describe the reported interaction experience.

We define learning gain as post-test percentage accuracy minus pre-test percentage accuracy, averaging sessions within each condition before calculating a participant's contrast:
\begin{equation}
L_{pc}=\overline{\mathrm{Post}}_{pc}-\overline{\mathrm{Pre}}_{pc},\qquad
D_p=L_{p,\mathrm{interactive}}-L_{p,\mathrm{noninteractive}}.
\end{equation}
Under this definition, gains are measured in percentage points rather than relative percent improvement. Within-participant contrasts compare the gain in the interactive condition with the gain in the noninteractive condition. {The reported mean learning gains of 47.9 percentage points with interaction and 32.1 without interaction, a difference of 15.8 points, summarize the seven participants who passed the attention test. Figure~\ref{fig:preliminary-study} in the main text reports this seven-participant learning-gain sample; its experience-rating panel contains four respondents. These summaries should not be interpreted as all-nine estimates.}

\subsection{{Formative-study interface}}
\label{app:study-interface}
\begin{figure}[H]
\centering
\includegraphics[width=0.98\linewidth,height=0.72\textheight,keepaspectratio]{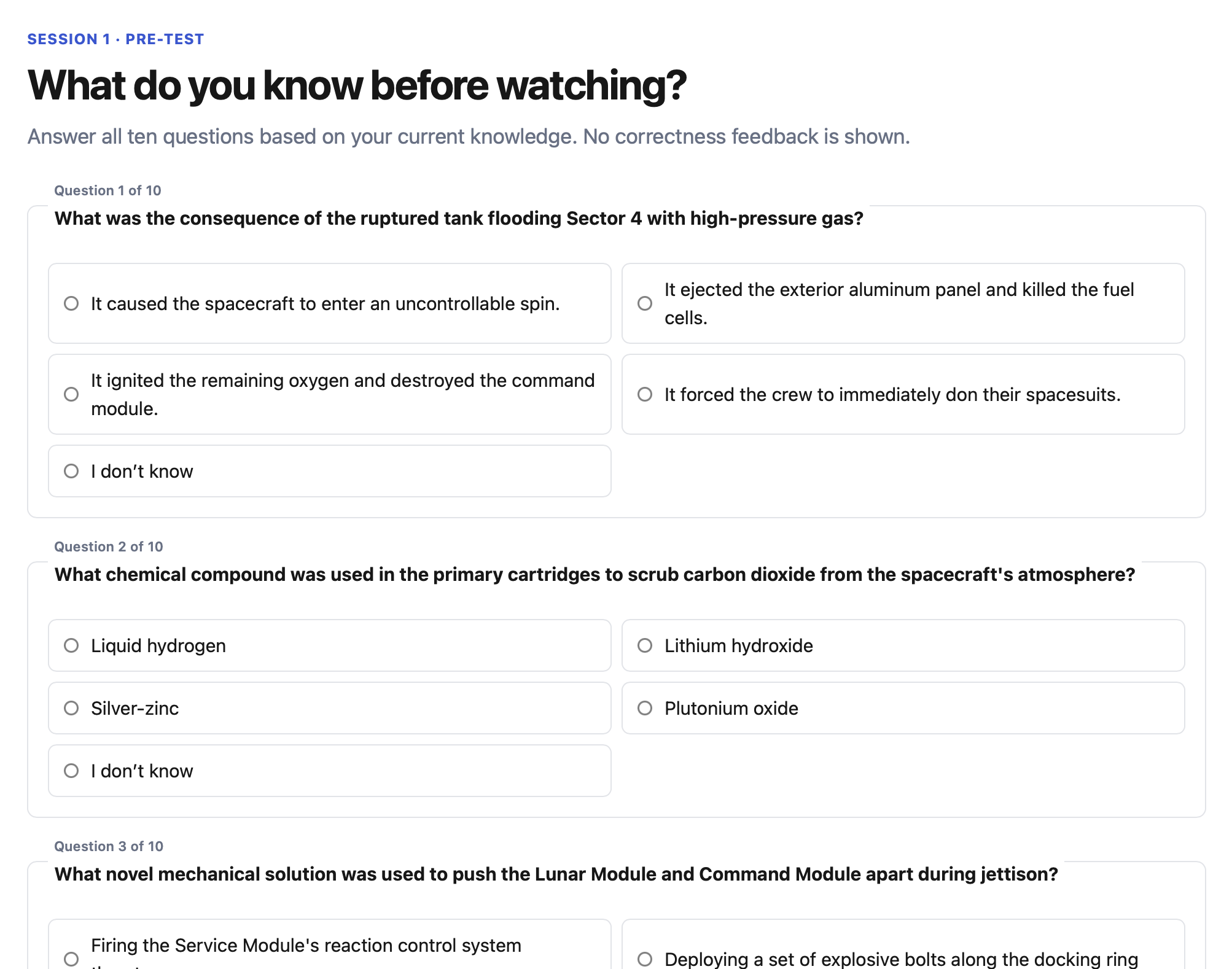}
\caption{{\textbf{Pre-test interface.} The pre-test screen asks participants to answer ten questions using their current knowledge, without correctness feedback. Each visible question includes an ``I don't know'' option.}}
\label{fig:study-pretest}
\end{figure}

\begin{figure}[H]
\centering
\includegraphics[width=0.98\linewidth,height=0.72\textheight,keepaspectratio]{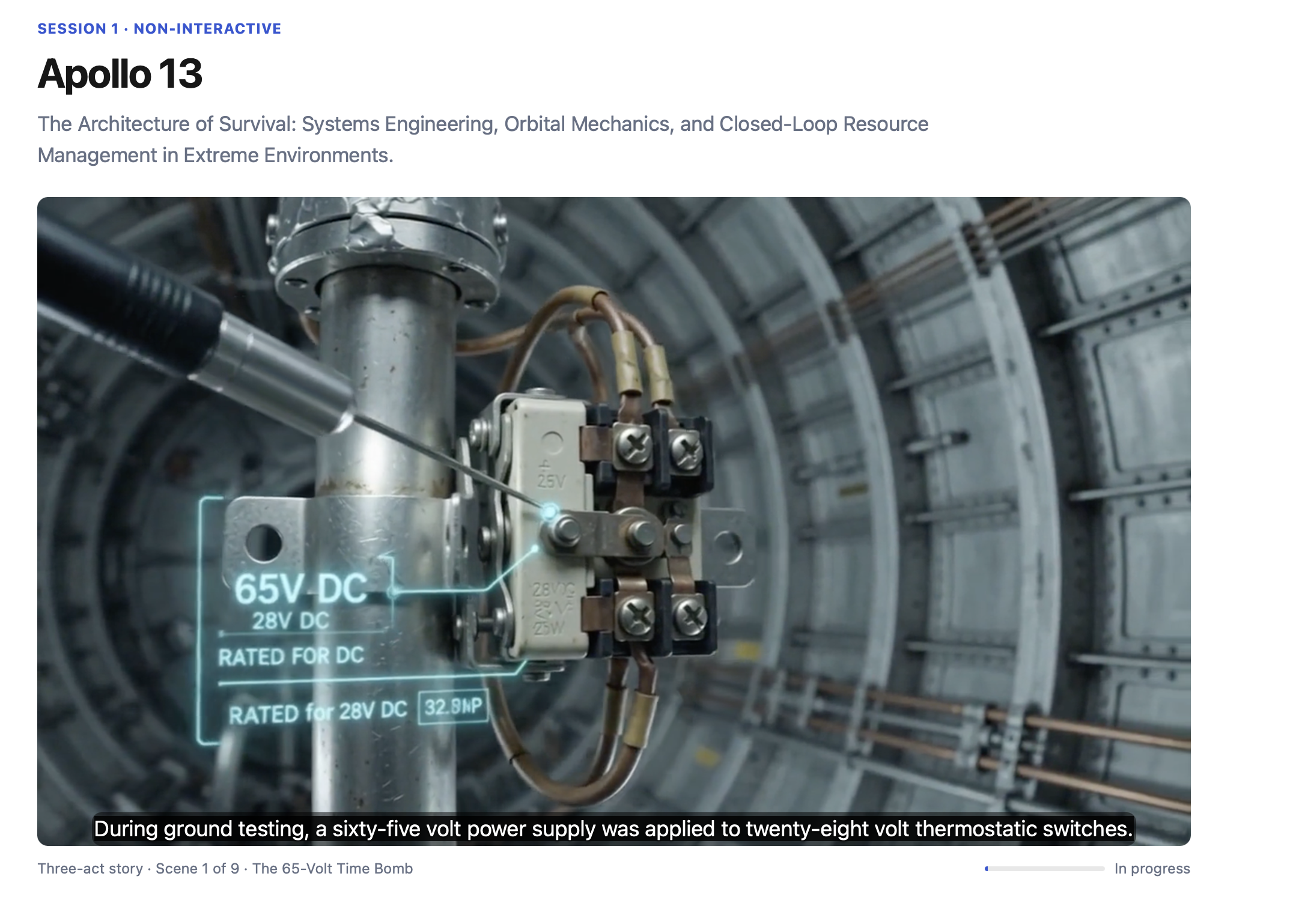}
\caption{{\textbf{Noninteractive viewing interface.} The noninteractive screen presents the video, subtitles, scene title, and viewing progress. The screenshot shows an Apollo 13 example.}}
\label{fig:study-noninteractive}
\end{figure}

\begin{figure}[H]
\centering
\includegraphics[width=0.98\linewidth,height=0.72\textheight,keepaspectratio]{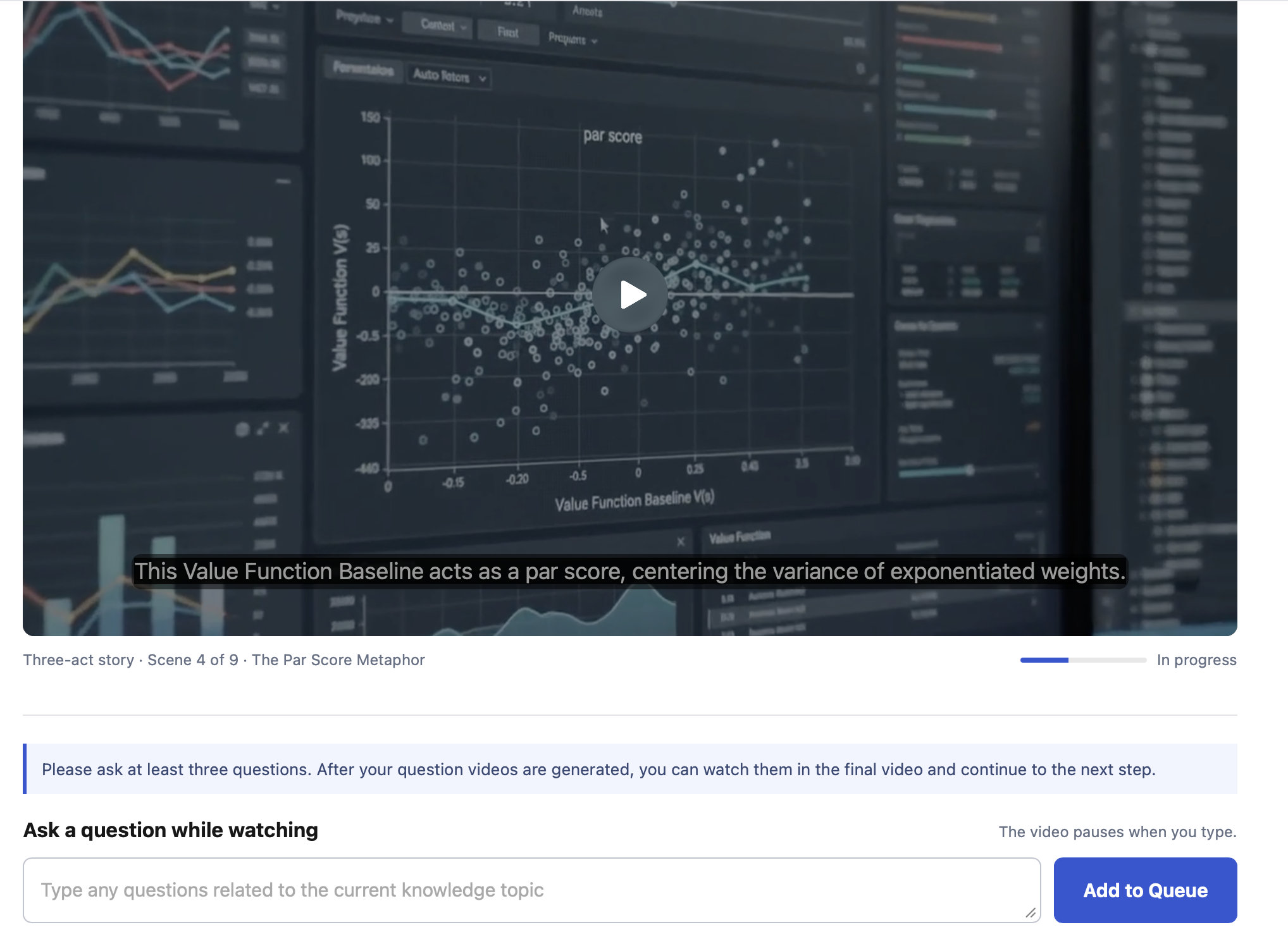}
\caption{{\textbf{Interactive viewing interface.} The interactive screen provides a question field and an ``Add to Queue'' control below the video. The displayed instructions ask participants to submit at least three questions and state that playback pauses while they type.}}
\label{fig:study-interactive}
\end{figure}

{Figures~\ref{fig:study-pretest}--\ref{fig:study-interactive} document the interface used in the formative learning study: the pre-test, noninteractive viewing, and interactive question-entry screens. These screenshots illustrate the participant-facing workflow; the attention-test exclusion criterion is described above.}

\subsection{Human validation of automatic evaluation}
\label{app:human-validation}

We additionally conduct a human validation study to examine whether the
automatic evaluators produce judgments consistent with human assessment.
This analysis is separate from the formative learning study and focuses
only on agreement in output-level quality judgments. We sample 50
generated outputs spanning both grounding settings, multiple generation
methods, and a range of automatic evaluation scores. Each output is
independently rated by three human annotators using the same metric
definitions and ordinal rating scale as the automatic evaluation.
Agreement is measured using Krippendorff's $\alpha$, treating the
\paperchange{median of the three human ratings} as the human reference when comparing against
automatic evaluators.

After the evaluation prompt calibration, human annotators exhibit
strong inter-rater agreement across all five metrics
($\alpha=0.78$--$0.86$). Both automatic evaluators also align closely
with human judgments, with Gemini reaching $\alpha=0.75$--$0.83$ and
Claude reaching $\alpha=0.64$--$0.76$. The results suggest that the
automatic evaluation provides a reliable approximation of human
judgment while enabling evaluation at substantially larger scale.

\begin{table}[htbp]
\centering
\small
\renewcommand{\arraystretch}{1.15}
\begin{tabular}{p{0.35\linewidth}p{0.57\linewidth}}
\toprule
\textbf{Field} & \textbf{Study details} \\
\midrule
Agreement statistic and scale
&
Krippendorff's $\alpha$ computed over the same ordinal rating scale
used by the automatic evaluators. \\[5pt]

Output sampling and coverage
&
50 generated outputs sampled across narrative and knowledge grounding,
multiple generation methods and backbones, and different automatic-score
ranges to obtain broad coverage of output quality. \\[5pt]

Rater recruitment and training
&
Three human annotators familiar with multimodal content evaluation.
Annotators first reviewed the metric definitions, rubric, and several
worked examples before independently completing the validation set. \\[5pt]

Ratings per output
&
Each output received three independent human ratings for SF, SC, IF,
VC, and VQ. \\[5pt]

Blinding and presentation order
&
Annotators were not shown the generation method, model identity, or
automatic evaluator scores. Outputs were presented in independently
randomized order. \\[5pt]

Disagreement handling
&
Ratings were collected independently without discussion.
For human--automatic agreement, the median human rating across the
three annotators was used as the reference judgment. \\[5pt]

Rubric calibration / validation separation
&
A small calibration set was used only to familiarize annotators with
the rubric and was excluded from the reported validation results.
The final validation set was rated independently after calibration. \\
\bottomrule
\end{tabular}
\caption{Protocol used for human validation of the automatic evaluation.}
\label{tab:human-protocol}
\end{table}

\subsection{Prompt and implementation availability}
\label{app:prompt-availability}
The prompt manifest distinguishes shared extraction templates, two-scene planning and refinement, the separate one-scene implementation, optional audio grounding, and final evaluation. Appendix~\ref{app:historical-prompts} reproduces eight extraction and two-scene planning/review templates.

\section{Prompt Templates}
\label{app:historical-prompts}
The following templates specify state extraction and two-scene planning and repair. Braced fields denote runtime substitutions, and output-schema names refer to the structured implementation interfaces. The first four templates extract and align the grounding context; the final four specify the 48-second narrative, shot expansion, review, and repair. Prompt instructions express intended behavior rather than guarantees of generated-output correctness.

\subsection{Atomic source facts}
\begin{Verbatim}[breaklines=true,fontsize=\footnotesize]
Extract atomic ordered facts from the original source for source-grounded video continuation. Preserve exact supporting quotes. For narrative text, include events, actions, speech acts, states, relationships, locations, objects, and causal dependencies. For knowledge text, include claims, mechanisms, causes, effects, examples, comparisons, and explanatory dependencies. Assign F001, F002, ... in source order. Do not add external knowledge. Return only SourceFacts JSON.

GROUNDING: {grounding_type}
SOURCE:
{source_text}
\end{Verbatim}

\subsection{Observed prefix ledger}
\begin{Verbatim}[breaklines=true,fontsize=\footnotesize]
Inspect the attached complete video prefix, especially its final seconds. Build a conservative state ledger for planning the immediate continuation. Record visible identities, appearance, location, object and character states, spatial relationships, recent and unresolved events, style, composition, and screen direction. Distinguish persistent from momentary state. Do not infer unseen outcomes. Put uncertain claims only in excluded_uncertain_claims. Return only PrefixStateLedger JSON.

STORY SUMMARY: {story_summary}
AVAILABLE ASSET IDENTITIES: {assets}
\end{Verbatim}

\subsection{Source--prefix alignment}
\begin{Verbatim}[breaklines=true,fontsize=\footnotesize]
Align atomic source facts to the prefix ledger. Partition every fact ID exactly once into shown, partially shown, unseen, or uncertain. "Shown" means semantically realized in the prefix, not merely mentioned in metadata. "Partially shown" is reserved for an event actively underway at the cut. Explain the continuation boundary conservatively. Return only SourceAlignment JSON.

INTERACTION SCENE LABEL: {interaction_after_scene}
FACTS: {source_facts}
PREFIX LEDGER: {prefix_ledger}
\end{Verbatim}

\subsection{Instruction requirements}
\begin{Verbatim}[breaklines=true,fontsize=\footnotesize]
Decompose the viewer instruction into atomic assessable requirements and map each semantic requirement only to relevant unseen or partially shown source facts. Presentation preferences may have no fact ID. Report requests contradicted or unsupported by the source in unsupported_requests. Do not map to already shown facts. Return only InstructionRequirements JSON.

GROUNDING: {grounding_type}
INSTRUCTION: {instruction}
FACTS: {source_facts}
ALIGNMENT: {source_alignment}
\end{Verbatim}

\subsection{Two-scene planning}
\begin{Verbatim}[breaklines=true,fontsize=\footnotesize]
You are a source-grounded narrative director. Expand the frozen interactive-video continuation into exactly two coherent scenes, each 24 seconds long. Each scene will later be divided into three 8-second shots.

GROUNDING: {grounding_type}
PREFIX SUMMARY: {story_summary}
ORIGINAL SOURCE:
{source_text}

VIEWER INSTRUCTION: {instruction}

FROZEN AGENT SCENE:
{scene_script}

AVAILABLE ASSETS:
{assets}

STRICT RULES
1. Return exactly two ordered scenes with indexes 1 and 2 and total_duration_seconds=48.
2. Both scenes together must fulfill the viewer instruction. Scene 1 begins fulfillment immediately; scene 2 advances or resolves it without replay.
3. Use only facts supported by the original source and only source_fact_ids, requirement_ids, and selected_asset_ids available in the frozen agent scene and asset list.
4. Scene 1 starts exactly from the attached prefix terminal state. Scene 2 must follow causally from scene 1, with an explicit boundary classified as exact_continuation, identity_composition_anchor, or intentional_jump.
5. Preserve identities, clothing, object state, relationships, environment logic, screen direction, and all persistent state unless the visible action explicitly changes them.
6. Each scene has a clear narrative purpose, a compact set of visible actions, and a precise initial_state/final_state contract. Scene 2 final_state must equal the frozen agent scene final_state.
7. Speech must be source-supported, concise enough for a single 8-second shot, and must not invent quotations or claims. Use narrator when sourced exposition is needed. A silent scene has no narration or dialogue.
8. Use exactly one exclusive audio mode per scene. Character dialogue contains exactly one line and must use a selected character asset as speaker.
9. Do not add unsupported characters, props, locations, facts, dialogue, outcomes, captions, or readable text.
10. Return only structured JSON conforming exactly to TwoSceneNarrativePlan.
\end{Verbatim}

\subsection{Six-shot expansion}
\begin{Verbatim}[breaklines=true,fontsize=\footnotesize]
You are a master cinematographer and continuity director. Turn the frozen two-scene narrative below into exactly six generation-ready shots: three ordered 8-second shots per scene.

PREFIX SUMMARY: {story_summary}
VIEWER INSTRUCTION: {instruction}
TWO-SCENE NARRATIVE:
{two_scene_plan}

AVAILABLE ASSETS:
{assets}

STRICT RULES
1. Return two SceneShotPlans. Scene 1 uses 0-8, 8-16, 16-24 seconds. Scene 2 uses 24-32, 32-40, 40-48 seconds. Every shot duration is exactly 8 seconds.
2. Each scene has exactly three shots indexed 1, 2, 3. Each shot has one dominant narrative beat, one focal action, readable composition, precise camera movement, and a rich standalone video_prompt.
3. Scene 1 shot 1 uses prefix_continuation. Within a scene, use exact_continuation for continuous motion/composition or hard_cut for a motivated camera-angle change.
4. Scene 2 shot 1 follows the supplied SceneBoundary. Use exact_continuation for an exact boundary; otherwise use hard_cut while preserving the boundary's required persistent state.
5. Preserve identities, clothing, props, environment, lighting logic, screen direction, spatial relationships, scale, and causal state across shots and across the scene boundary.
6. Use only fact IDs, requirement IDs, and asset IDs assigned to the corresponding narrative scene. Do not add facts, dialogue, characters, props, locations, or outcomes.
7. Distribute each scene's visible actions across its three shots without replay. The last shot of each scene must reach that scene's frozen final_state.
8. For a narrator or dialogue scene, exactly one of its three shots uses scene_audio; the other two use ambient_only. For a silent scene, all three use ambient_only.
9. In ambient_only shots, nobody speaks, talks, addresses an audience, narrates, or lip-syncs; prompts specify closed, non-speaking mouths.
10. Each video_prompt describes only its own 8-second shot and prohibits subtitles, captions, title cards, logos, watermarks, and unsupported readable text.
11. Keep human faces natural and anatomically normal with restrained expressions. Avoid identity, costume, object, and scale drift, extra limbs, and implausible motion.
12. Return only structured JSON conforming exactly to TwoSceneShotPlan.
\end{Verbatim}

\subsection{Plan review}
\begin{Verbatim}[breaklines=true,fontsize=\footnotesize]
Audit the complete six-shot continuation using only the public source, instruction,
actual prefix and extracted public states. Never use external biographical or historical knowledge.
Check that required visual actions actually happen instead of being narrated or replaced by decorative motion.
Every shot must advance a distinct part of the requested continuation. Avoid spending several shots on a static presentation.
Check initial-to-final action states, motivated transitions, persistent identity/wardrobe/object geometry,
screen direction and literal material style. A paper-cutout scene must not become glossy realistic 3D.
Check exact speech fits its eight-second shot at a natural pace (usually at most 18–20 words),
and that any audio-only shot is intentional. Do not demand on-screen text unless the instruction requires it.
Distinguish scene changes from identity drift. Report concrete defects with affected shot numbers.
Accept only if no material defect is found. These are generation checks, not private evaluation labels.
\end{Verbatim}

\subsection{Plan repair}
\begin{Verbatim}[breaklines=true,fontsize=\footnotesize]
Repair the six-shot Plan using the public evidence and listed defects. Preserve its 48-second,
two-scene structure and every source-grounded requirement.
Mandatory structural fields: shot_id exactly [1,2,3,4,5,6]; scene_id exactly [1,1,1,2,2,2];
duration exactly 8 for every shot. Shot 1 transition MUST be prefix_continuation and must visually
start from the actual prefix terminal state. No other shot may use prefix_continuation.
If the story needs a new setting, establish that change with a hard_cut in shot 2 or later;
never label shot 1 hard_cut. These fixed fields override any informal editing suggestion.
Use only listed asset IDs and at most the
supplied reference limit per shot. State literal visual materials and stable identifying attributes
in each standalone shot prompt. Give each shot one executable focal action with a visible start and end.
Make later shots continue the achieved state without replay. Use a motivated hard_cut when the setting
changes; do not ask for an impossible continuous transformation of location or identity. Keep speech
brief and source-supported; never invent facts to increase detail. Return the complete Plan.
\end{Verbatim}

\end{document}